%% file: main.tex
\documentclass[10pt,twocolumn,letterpaper]{article}

\usepackage[pagenumbers]{cvpr} % To force page numbers, e.g. for an arXiv version

\input{preamble}
\usepackage{amssymb}  

\usepackage{pifont}
\newcommand{\cmark}{\ding{51}}
\newcommand{\xmark}{\ding{55}}

\definecolor{cvprblue}{rgb}{0.21,0.49,0.74}
\usepackage[pagebackref,breaklinks,colorlinks,allcolors=cvprblue]{hyperref}

\newcommand{\methodname}{BiMotion\xspace}

\def\paperID{36025} % *** Enter the Paper ID here
\def\confName{CVPR}
\def\confYear{2026}

\title{\methodname: B-spline Motion for Text-guided Dynamic 3D Character Generation}
\author{
Miaowei Wang$^{1}$ \quad
Qingxuan Yan$^{2}$ \quad
Zhi Cao$^{3}$ \quad
Yayuan Li$^{3}$ \quad \\
Oisin Mac Aodha$^{1}$ \quad
Jason J. Corso$^{3,4}$ \quad
Amir Vaxman$^{1}$ \\
$^{1}$ University of Edinburgh \quad
$^{2}$ Cornell University \quad
$^{3}$ University of Michigan \quad
$^{4}$ Voxel51
}

\usepackage{amsmath}

\usepackage{cvpr}              % To produce the CAMERA-READY version
\usepackage{xcolor}
\definecolor{myblue}{HTML}{367DBD}
\usepackage{siunitx}
\usepackage{amsmath}
\usepackage{amsmath}
 \usepackage{hyperref}
\usepackage{balance}
\usepackage{algorithm}
\usepackage{algorithmic}
\usepackage{makecell}

\input{preamble}
\def\confName{CVPR}
\def\confYear{2026}

\begin{document}

\twocolumn[{%
\renewcommand\twocolumn[1][]{#1}%
\maketitle
\begin{center}
    \centering
    \captionsetup{type=figure}
\vspace{-7mm}
\includegraphics[width=1\textwidth]{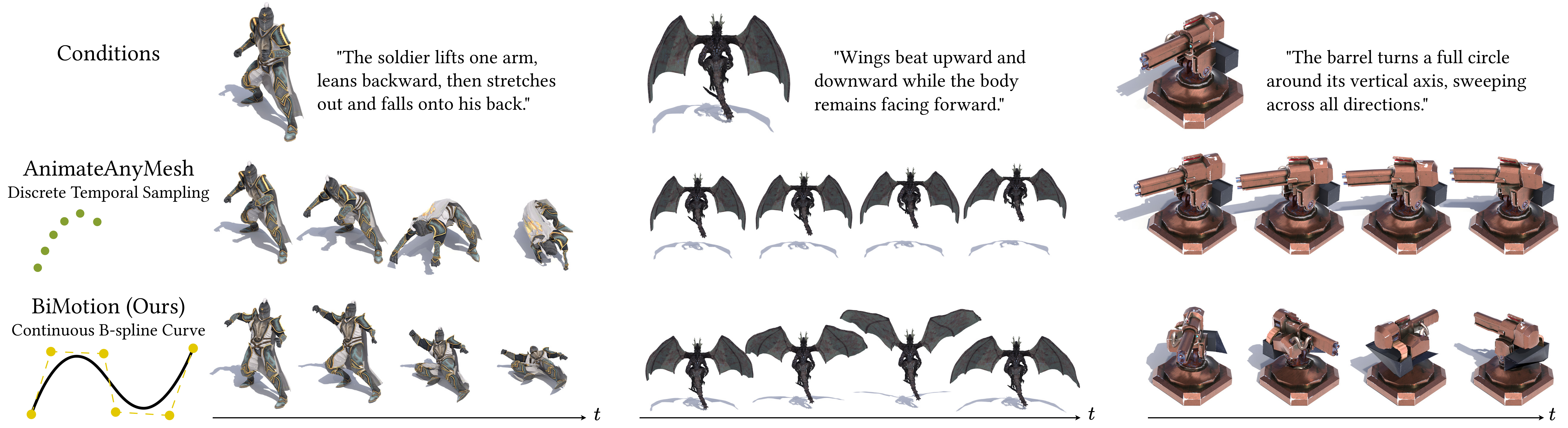}
    \vspace{-7mm}
    \captionof{figure}{We propose \textbf{\methodname}, a fast, feed-forward B-spline–based method for dynamic 3D character generation. 
    It produces continuous, high-quality expressive motion trajectories aligned with rich textual prompts, outperforming discrete temporal sampling-based methods such as AnimateAnyMesh~\cite{wu2025animateanymesh} under the same fixed-input constraint. 
    See our project page for full motion dynamics.}
\vspace{-1.5mm}
\label{fig:teaser}
\end{center}
}]

\input{sec/0_abstract}    
\input{sec/1_intro}
\input{sec/2_related_works}

\input{sec/3_methodology}
\input{sec/4_experiment}
\input{sec/6_conclusion}
\section{Acknowledgement}
This research was funded, in part, by the U.S. Government under ARPA-H contract 1AY2AX000062. The views and conclusions contained in this document are those of the authors and should not be interpreted as representing the official policies, either expressed or implied, of the U.S. Government.
\input{Suppl}

\clearpage
\newpage
{
    \small
    \bibliographystyle{ieeenat_fullname}
    \bibliography{main}
}
\end{document}

%% file: preamble.tex
\PassOptionsToPackage{numbers}{natbib}
\usepackage{natbib}
\usepackage{float}
\usepackage{multirow}
\usepackage{booktabs}
\usepackage{siunitx}
\usepackage{amsmath}

\usepackage[table]{xcolor}

\definecolor{LightCyan}{rgb}{0.88,1,1}

\definecolor{myteal}{RGB}{72,255,204}      % MediumTurquoise
\definecolor{myblue}{RGB}{30,144,255}      % DodgerBlue

%% file: sec/0_abstract.tex
\begin{abstract} 
Text-guided dynamic 3D character generation has advanced rapidly, yet producing high-quality motion that faithfully reflects rich textual descriptions remains challenging. Existing methods tend to generate limited sub-actions or incoherent motion due to fixed-length temporal inputs and  discrete frame-wise representations that fail to capture rich motion semantics.
We address these limitations by representing motion with continuous differentiable B-spline curves, enabling more effective motion generation without modifying the capabilities of the underlying generative model. Specifically, our  closed-form, Laplacian-regularized B-spline solver efficiently compresses variable-length motion sequences into compact representations with a fixed number of control points.  
Further, we introduce a normal-fusion strategy for input shape adherence along with correspondence-aware and local-rigidity losses for motion-restoration quality. 
To train our model, we collate BIMO, a new dataset containing diverse variable-length 3D motion sequences with rich, high-quality text annotations.  Extensive evaluations show that our feed-forward framework \methodname generates more expressive, higher-quality, and better prompt-aligned motions than existing state-of-the-art methods, while also achieving faster generation. Our project page is at: \url{https://wangmiaowei.github.io/BiMotion.github.io/}.
\end{abstract}

%% file: sec/1_intro.tex
\section{Introduction}
\label{sec:intro}
Recent advances in content creation have shifted from static 3D generation \cite{zhao2025hunyuan3d,xiang2025structured,chen2025partgen,yang2025hash3d} toward dynamic 3D generation \cite{miao2025advances,wu2025cat4d,wang20254real}, driven by growing demands in gaming, film, and educational applications.
Rather than simultaneously generating both geometry and motion \cite{zhang2024dnfunconditional4dgeneration,jiang2024animate3d,ren2024l4gm,huangmvtokenflow,huang2025animax}, by decoupling motion generation from shape synthesis, recent methods \cite{wu2025animateanymesh,CanFields2025,zhang2025gaussian,rahamim2024bringingobjectslife4d} focus solely on animating a given initial shape by leveraging existing 3D assets or mature generation models \cite{zhao2025hunyuan3d,10684147}. 
This decoupling paradigm preserves subject consistency and enables direct use of existing 3D content libraries \cite{sketchfab_free_models,free3d_models,cgtrader_free_models,turbosquid_free_models,clara_library}.

%\AV{In the following, you'll need to forward reference examples in your paper that demonstrate this}
However, current feed-forward motion generation methods \cite{wu2025animateanymesh,zhang2024dnfunconditional4dgeneration,zhang2025gaussian} face a critical limitation: they struggle to produce expressive, semantically complete motion that faithfully matches the user description (see \cref{fig:teaser}). 
Existing methods typically employ VAE–latent diffusion models \cite{rombach2022high}, with fixed-size inputs, forcing them to either crop motion sequences into short fragments \cite{liang2024diffusion4d,wu2025animateanymesh,zhang2025gaussian} or uniformly downsample them \cite{zhang2024dnfunconditional4dgeneration,li2025puppetmaster}. 
Cropping breaks motion completeness, capturing only isolated sub-actions like ``rotate to the right'' instead of the full user prompt (\cref{fig:teaser}, Right). 
Temporal downsampling yields non-smooth, jittery results \cite{zhang2024dnfunconditional4dgeneration}. 
As a result, rich motion semantics present in variable-length datasets \cite{li20214dcomplete,deitke2023objaverse,deitke2023objaverseXL,CAPE:CVPR:20}, which contain motion spanning tens  to hundreds of frames,  are not captured.

The fundamental challenge is this: \emph{how can variable-length, semantically complete motions be represented within capacity-limited models that require fixed-size inputs?} Motion is inherently continuous, with frame count merely reflecting temporal sampling and semantic meaning (\eg ``a dragon flapping its wings'') is unchanged whether captured at 24 or 120 frames.  
Discrete frame-wise representation  is hence a key bottleneck. 
Therefore, what is needed is a continuous, compact parameterization that preserves full motion semantics regardless of sequence length.

We introduce \textbf{\methodname}, a B-spline-based motion representation that directly addresses this challenge. B-splines \cite{bonet1997nonlinear,haron2012parameterization} model continuous trajectories using a compact set of control points, providing exactly the parameterization we need. 
B-splines offer three properties ideal for representing motion: (1) continuity and differentiability for natural trajectories; 
(2) local controllability, where modifying one control point only affects its neighborhood; 
and (3) temporal reparameterization, allowing sampling at any time along a trajectory defined by control points.
We further develop a closed-form, Laplacian-regularized B-spline solver that efficiently represents variable-length motion into fixed-size control point sets, preserving global motion semantics while maintaining continuous motion evolution.

Integrating B-spline control points into VAE–latent diffusion models as inputs requires rethinking how motion information is embedded.
We propose a novel multi-level control-point embedding tailored for B-spline representation, which significantly outperforms standard frequency-based positional encodings~\cite{mildenhall2021nerf} in motion reconstruction (\cref{fig:CPE_ablation}). 
We also introduce a normal fusion strategy that enhances the VAE’s geometric awareness, effectively distinguishing closely positioned motion parts while remaining robust and efficient, unlike mesh-connectivity methods~\cite{wu2025animateanymesh}, which become unstable when scaled to many vertices. 
Our B-spline formulation enables projecting predicted control points back onto motion trajectories, supervised by a correspondence loss aligning with raw trajectory differences and a local rigidity loss preserving shape identity.
To train \methodname, we construct \textbf{BIMO}, a dataset comprising $\sim$39K sequences with control points, raw differences trajectory, and high-quality textual descriptions generated through a hybrid of human and automatic annotation with iterative feedback to ensure consistency and accuracy. 
% Our experiments show that the proposed B-spline–based approach generates more expressive and plausible motions with higher fidelity and faster runtime than existing methods.

In summary, our main contributions are:
\begin{itemize}
\item \methodname, a B-spline–based feed-forward framework that generates faster, more plausible, expressive, and semantically complete text-guided motion from an initial 3D shape than state-of-the-art methods (SOTAs).
\item A specialized VAE with novel control-point embedding, normal fusion, and correspondence and rigidity losses tailored to B-spline motion representation.
\item BIMO, a general dataset of $\sim$39K high-quality annotated motions represented by B-spline curves, and extensive experiments validating our approach's effectiveness.
\end{itemize}

%% file: sec/2_related_works.tex
\begin{figure*}[!htb]
    \centering
     \includegraphics[width=1\linewidth]{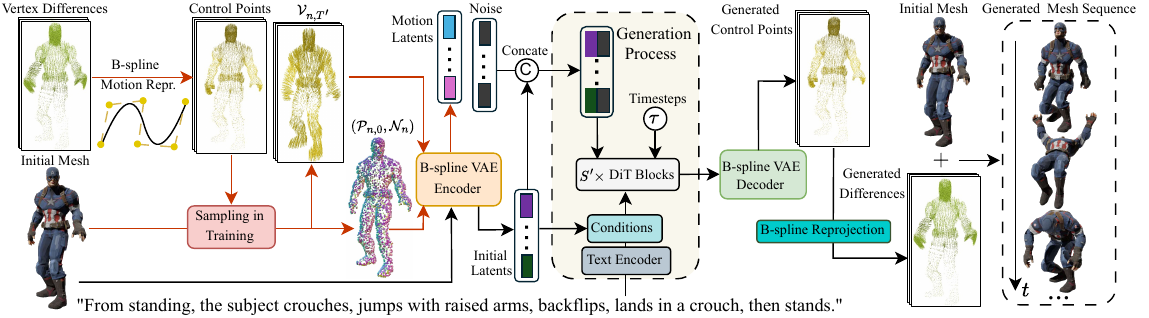}
\vspace{-6mm}
\caption{\textbf{Overview.} \methodname uses a B-spline representation for motion generation. 
During training (red arrow), vertex differences are converted into control points and encoded into motion latents. 
During inference (black arrow), the initial mesh and the text prompt generate motion latents that are decoded into control points and converted into the generated mesh sequence via B-spline reprojection.}
\vspace{-4mm}
\label{fig:overview}
\end{figure*}

\vspace{-1mm}
\section{Related Work}
\label{sec:related_works}
\vspace{-1mm}
\noindent \textbf{Static and Dynamic 3D Generation.}
Static and dynamic 3D generation methods follow similar technical trajectories. 
initial static approaches optimize shapes using image-generation models via Score Distillation Sampling (SDS) and its variants \cite{tang2023stable,wang2023prolificdreamer,poole2023dreamfusion}. 
Dynamic 3D methods have used the same paradigm by extracting motion priors from video-generation models through video-based SDS variants \cite{bah20244dfy,li2025articulated,yu20244real}. 
As the Janus (multi-face) problem became evident \cite{wei2024dreamvideo,hong2023debiasing}, subsequent works shifted to multi-view image diffusion \cite{huang2025mv,liu2023one} and multi-view video diffusion \cite{zhang20244diffusion,xie2025videopanda,qin2025distilling,jiang2024animate3d} to achieve view-consistent static \cite{shimvdream,gao2024cat3d,liu2024one,MVD2} and dynamic 3D generation \cite{ICLR2025_4c2eb991,wu2025cat4d,ren2024l4gm,huangmvtokenflow,huang2025animax,gao2025charactershot}. 
To mitigate geometry collapse and enable faster feed-forward generation, recent methods train directly on 3D assets instead of using multi-view projections \cite{hanflex3d,zhao2023michelangelo,xiong2025octfusion,zhang20233dshape2vecset,zhao2025hunyuan3d}. 
The same paradigm has been applied to dynamic 3D generation \cite{wu2025animateanymesh,zhang2025gaussian}. 
These models typically employ VAE-latent diffusion architectures that encode dynamic 3D assets into fixed-size latent spaces, which restricts variable-range motion awareness. 
Autoregressive models \cite{tian2024visual,zhu2025ar4d} support variable-length sequences but are computationally expensive and struggle to generalize beyond human skeletons \cite{xiao2025motionstreamer,pinyoanuntapong2024bamm,tuautoregressive} for arbitrary character motion generation. 
To strike a balance, our work focuses on enabling a fixed-capacity model to support diverse, variable-length motions.

\vspace{1mm}
\noindent \textbf{Conditional Motion Generation.} 
Letting a user specify the desired motion---conditional generation---is essential for creating controllable, usable dynamic 3D assets in creative applications.  Unconditional motion generation \cite{zhang2024dnfunconditional4dgeneration,raab2023modi,bjorkstrand2025unconditional}, however, does not support this rich need.
Many works \cite{chen2025v2m4,zhang2025gaussian,wangfused,yang2025not,wang2024vidu4d} target video-to-dynamic-3D generation because videos provide richer motion cues than text alone \cite{wu2025animateanymesh,chen2024ct4d,dai2025textmesh4d,Shao_2024_CVPR,hong2022avatarclip}. However, these methods are biased toward reconstruction from videos: generated motions can be sensitive to camera, content, and lighting changes \cite{chen2025v2m4,zhang2025gaussian,yao2025sv4d}, with quality largely dependent on the source video \cite{yin20234dgen}.
Some methods incorporate user-specified trajectories for finer control \cite{bahmani2024tc4d,mao2025dreamdrive,wang2024occsora,li2025puppet}. 
Dynamic 3D generation can also be conditioned on the first and last frames \cite{sang2025twosquared,sang20254deform,cosmo2020limp}, though this is closer to motion interpolation. 
Following \cite{wu2025animateanymesh}, we condition motion generation on text and an initial 3D mesh, while our method remains adaptable to other conditioning inputs.
%but we are  agnostic to the precise form of conditioning provided.

\vspace{1mm}
\noindent\textbf{Motion Representation.} In current dynamic 3D character pipelines, rigging, including skeletons and skinning, is essential for controlling shape deformation. 
Many works automatically estimate rigging \cite{guo2025make,deng2025anymate,zhang2025one,songpuppeteer}, which can drive motion generation by optimizing the skeleton using video \cite{songpuppeteer,li2025articulated,li2025articulated}. Skeleton-based feed-forward approaches are commonly used in human motion generation \cite{chu2025humanrig,khani2025unimogen,genmo2025}, but they typically require sequences padded to a predefined maximum  length. 
\citet{10.1145/3721238.3730621} perform animal skeleton-driven motion generation, but still require a fixed number of skeleton joints, \citet{zhang2025physrig} incorporate a physics engine to enhance elastic effects, and \citet{he2025category,zhang2023tapmo} learn rigging-like sparse handles from datasets to facilitate user-driven shape deformation. 
AnimateAnyMesh~\cite{wu2025animateanymesh} learns motion tokens by sampling vertices, offering a potential route for universal feed-forward motion generation. 
However, these representations focus on shape deformation, with motion sequences still composed naively on a frame-by-frame basis.  In contrast, we represent variable-length motion using B-spline curves. Splines \cite{10.1145/54852.378512} have long been used in geometric modeling \cite{bajaj2017splines,ge2023shape} and, more recently, in dynamic scene reconstruction~\cite{song2025spline,park2025splinegs}. To our knowledge, we are the first to employ spline-based representations for motion generation.

%% file: sec/3_methodology.tex
%\clearpage
\section{Methodology}
\label{sec:method}
\begin{figure*}[!htb]
    \centering
     \includegraphics[width=1\linewidth]{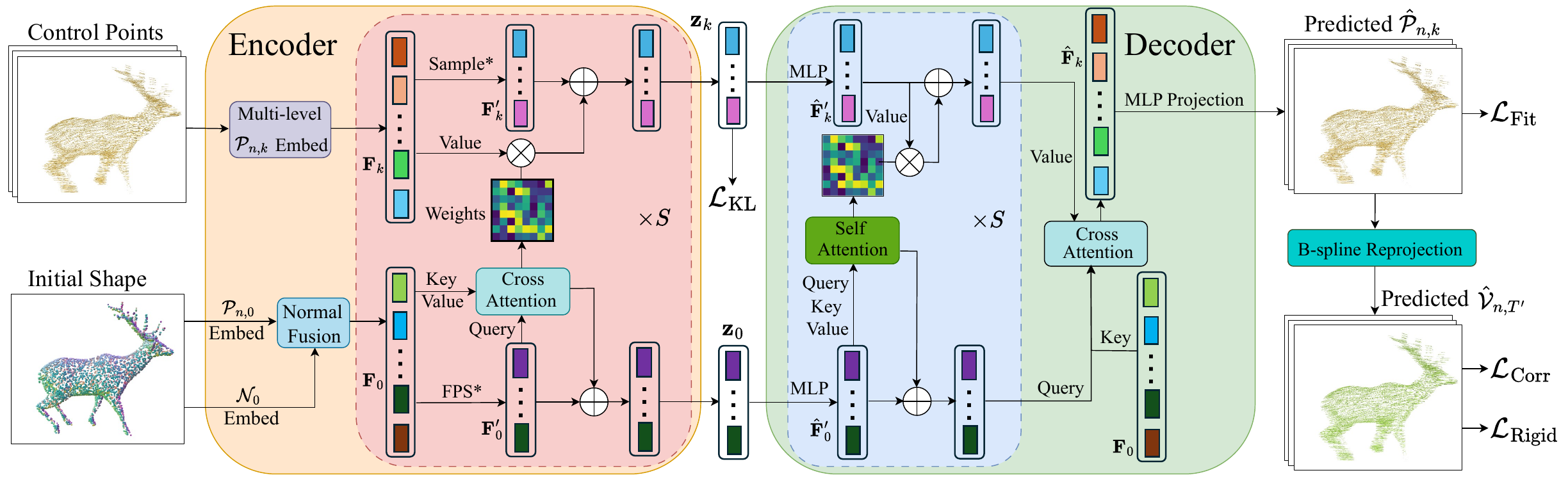}
\vspace{-6mm}
\caption{\textbf{B-spline VAE Pipeline.} Given the initial shape $(\mathcal{P}_{n,0}, \mathcal{N}_0)$ and control points $\mathcal{P}_{n,k}$, the \textit{Encoder} compresses them into latent codes $\mathbf{z}_0$ and $\mathbf{z}_{k}$. The \textit{Decoder} reconstructs the predicted control points $\hat{\mathcal{P}}_{n,k}$, which are then reprojected to point differences via the B-spline basis. Note, * indicates that $\mathbf{F}_{k}'$ uses the FPS-sampled indices of $\mathbf{F}_0'$, $\oplus$ denotes matrix addition, and $\otimes$ denotes matrix multiplication.}
\vspace{-4mm}
\label{fig:VAE_Pipeline}
\end{figure*}

\subsection{Problem Statement}
\label{subsec:problem}

\noindent Given a dataset of dynamic mesh sequences $\mathcal{D}=\{(M^{(i)}_{0:T},y^{(i)})\}_{i=1}^N$, where each sequence $M_{0:T}$ has variable length $T+1$, $y$ is a user provided textual prompt, and $N$ is the total number of sequences, our goal is to learn a $\theta$-parameterized conditional posterior
$
p_{\theta}(M_{1:T}\mid M_0,y)\approx p_{\mathcal{D}}(M_{1:T}\mid M_0,y)
$. 
At inference we wish to animate an initial mesh $M_0$ by sampling from $p_{\theta}(\cdot\mid M_0,y)$ conditioned on the user prompt $y$ with the learned parameters~$\theta$. 
The initial mesh $M_0=(\mathbf{V}_0,\mathbf{F}_0)$ contains vertex positions $\mathbf{V}_0\in\mathbb{R}^{n_v\times3}$ and shared triangular faces $\mathbf{F}_0\in\mathbb{R}^{n_f\times3}$  across all subsequent timesteps. 
To disentangle motion from shape, following \cite{zhang2025gaussian,wu2025animateanymesh}, we represent all subsequent frames by vertex-wise displacements
$\{\mathbf{V}_t = \mathbf{V}_0+\Delta \mathbf{V}_t\}_{t=1}^T$ via the  initial vertex positions $\mathbf{V}_0$ and a vertex-wise difference $\Delta \mathbf{V}_t$. We stack vertex differences $\{\Delta \mathbf{V}_t\}_{t=1}^T$ into $\mathcal{V}_{T} \in \mathbb{R}^{T \times n_v \times 3}$.

\subsection{Overview} We employ a VAE-latent diffusion  approach for motion generation. 
During training, \methodname (\cref{fig:overview}) first converts variable-length vertex differences $\mathcal{V}_T$ into compact control points by representing each  difference trajectory  independently using B-splines. 
These control points are uniformly densely sampled, then randomly downsampled and encoded by a B-spline VAE encoder in each iteration to obtain motion latents.
During inference (\ie motion generation), the model is conditioned on a text prompt $y$ and an initial mesh $M_0$, and transforms noise into motion latents. 
The VAE decoder then generates control points from these latents, which are reprojected into vertex differences of arbitrary length to generate the final 3D motion sequence. 
We describe the details of each stage below.

\subsection{Preliminary: B-spline Curves}
\label{subsec:bspline-prelim}
A 3D B-spline curve $\mathcal{C}(t)$ of degree $d$ with $k$ control points $\{\mathbf{p}_i \in \mathbb{R}^3\}_{i=0}^{k-1}$ is defined over the (time) parameter $t$~\cite{bartels1995introduction} as:
\begin{equation}
\mathcal{C}(t)=\sum_{i=0}^{k-1} \mathcal{N}_{i,d}(t)\,\mathbf{p}_i,\quad
t\in [u_d,\,u_{m-1-d}].
\label{eq:bspline}
\end{equation}
The functions $\mathcal{N}$ are local, with their support defined by a non-decreasing knot vector $\mathbf{u}=\{u_0,\dots,u_{m-1}\}$ satisfying $m = k + d + 1$. The basis functions $\{\mathcal{N}_{i,d}(t)\}_{i=0}^{k-1}\}$ are built by the Cox--de Boor recursion \cite{deboor1978practical}. 

A \emph{clamped} knot vector repeats the first and last knot values $d+1$ times, i.e., $u_0=\cdots=u_d$ and $u_{m-1-d}=\cdots=u_{m-1}$, enforcing endpoint interpolation (\cref{fig:teaser}, Left) 
$
\mathcal{C}(u_d)=\mathbf{p}_0, \mathcal{C}(u_{m-1-d})=\mathbf{p}_{k-1}.
$
With clamped ends and uniformly spaced interior knots, the curve attains $\mathrm{C}^{d-1}$ continuity at interior knots. Sampling the B-splines at timesteps $t_{1},\dots,t_{T}$, we express the sampled points of \cref{eq:bspline} as:
\begin{equation}
\mathcal{C}_{T}=\mathcal{B} _{T,k}\mathcal{P}_{k},
\end{equation}
with the B-spline basis matrix $\mathcal{B}_{T,k} \in \mathbb{R}^{T\times k}$ stacked by rows $\{\big[\mathcal{N}_{0,d}(t),\dots,\mathcal{N}_{k-1,d}(t)\big]\}_{t=1}^{T}$ and the control-point matrix as $\mathcal{P}_{k}=[\mathbf{p}_0^\top;\dots;\mathbf{p}_{k-1}^\top]\in\mathbb{R}^{k\times3}$.

\subsection{B-spline Motion Representation}
\label{subsec:bspline-compress}
\noindent \noindent To obtain a fixed-size motion representation for variable-$T$ sequence, the differences $\mathcal{V}_{T}$ are approximated over time by the commonly used uniform cubic ($d=3$) B-splines with stacked control points $\mathcal{P}_{k} \in \mathbb{R}^{k \times n_v \times 3}$, using least-squares fitting over temporal samples: $
\min_{\mathcal{P}_{k}} \ \| \mathcal{B}_{T,k} \mathcal{P}_{k} - \mathcal{V}_T \|_F^2$.

When the number of control points $k \le T$, the least-squares solution is unique. For short sequences where $k > T$, the system becomes underdetermined. To regularize this, we apply a Laplacian term that enforces natural transitions between adjacent control points. Let $\mathbf{L}\in\mathbb{R}^{k\times k}$ denote the discrete second-order difference operator. We then solve:
\begin{equation}
% \vspace{-1mm}
\label{eq:bspline-reg}
\min_{\mathcal{P}_{k}} \;\|\mathcal{B}_{T,k} \mathcal{P}_{k} - \mathcal{V}_T\|_F^2 \;+\; \mu \,\|\mathbf{L} \mathcal{P}_{k}\|_F^2,
% \vspace{-1mm}
\end{equation}
which admits the closed-form solution 
\cite{CALVETTI2000423}:
\begin{equation}
\label{eq:closed-form}
% \vspace{10mm}
\mathcal{P}_{k} = \big(\mathcal{B}_{T,k}^\top \mathcal{B}_{T,k} + \mu \, \mathbf{L}^\top \mathbf{L}\big)^{-1} \mathcal{B}_{T,k}^\top \, \mathcal{V}_{T}.
% \vspace{-1mm}
\end{equation}

We denote the Laplacian-regularized least-square operator as $
\mathcal{O}_{T,k} :=(\mathcal{B}_{T,k}^\top \mathcal{B}_{T,k} + \mu \, \mathbf{L}^\top \mathbf{L})^{-1} \mathcal{B}_{T,k}^\top$ (see ablation in \cref{fig:ablation_lambda}),
which is efficiently computed via Cholesky decomposition~\cite{pytorch_cholesky}, \ie under one second on a consumer-grade CPU for 200-frame mesh sequences with 50\text{K} vertices.

To train across the full spatial field and ensure robustness to varying mesh topologies (\cref{fig:topology_robust}), following DNF~\cite{zhang2024dnfunconditional4dgeneration}, we uniformly pre-sample a dense set of $N\!=\!200\text{K}$ points with normals $\mathcal{N}_N$ from the initial mesh $M_0$. The corresponding control points $\mathcal{P}_{N,k}$ of the B-splines $\mathcal{C}_N(t)$ and point differences $\mathcal{V}_{N,T}$ are efficiently barycentric interpolated from $\mathcal{P}_k$ and $\mathcal{V}_T$, respectively. The $\mathcal{V}_{N,T}$ are temporally $T'$-uniformly sampled to $\mathcal{V}_{N,T'}$ for later training.

\subsection{B-spline VAE}
\label{subsec:bsplinevae}
Following the above motion representation, we introduce a B-spline-based VAE architecture (see \cref{fig:VAE_Pipeline}) to spatially compress motion into compact latents for subsequent generative training. In each training iteration, we randomly downsample $n\!<\!N$ initial points $\mathcal{P}_{n,0}\!\in\!\mathbb{R}^{n\times3}$ with normals $\mathcal{N}_{n}\!\in\!(\mathbb{S}^2)^{n}$ from the dense set in \cref{subsec:bspline-compress}, and collect their corresponding control points $\mathcal{P}_{n,k}\!\in\!\mathbb{R}^{k\times n\times3}$ and differences trajectory $\mathcal{V}_{n,T'}\!\in\!\mathbb{R}^{T'\times n\times3}$. The initial shape and control points serve as inputs to the VAE. For the encoder, $\mathcal{P}_{n,0}$ provide absolute 3D coordinates, which are lifted using cosine positional encoding~\cite{mildenhall2021nerf} and then projected via an MLP into $c$-dimensional point features $\mathbf{F}_{\mathcal{P}_{0}}\!\in\!\mathbb{R}^{n\times c}$.

\vspace{1mm}
\noindent\textbf{Normal Fusion.}
The normals $\mathcal{N}$ provide complementary local geometric cues, particularly where nearby points are far intrinsically on the mesh \cite{yang2020cn,Ran_2022_CVPR}. We embed normals with a two-layer MLP to obtain $\mathbf{F}_{\mathcal{N}_0}\in\mathbb{R}^{n\times c}$ \cite{yin2024point}. The normals are fused with the point features $\mathbf{F}_{\mathcal{P}_0}$ using a per-point cosine-similarity weight $w\!\in\![0,1]^n$ along the feature dimension, yielding the initial-shape embedding:
\begin{equation}
\mathbf{F}_0 = \mathbf{F}_{\mathcal{P}_{0}} + \big(w(\mathbf{F}_{\mathcal{P}_0},\mathbf{F}_{\mathcal{N}_0})\odot\mathbf{1}_{c}\big)\odot\mathbf{F}_{\mathcal{N}_0},
\label{eq:spatial-embedding}
\end{equation}
where $\odot$ denotes element-wise multiplication.

\vspace{1mm}
\noindent\textbf{Control Point Embedding.} Since conventional frequency encodings fail to capture the full motion patterns of control points (\cref{fig:CPE_ablation}), we introduce a novel multi-level hierarchical embedding for $\mathcal{P}_{n,k}$, inspired by wavelet packet decomposition \cite{ting2008eeg}. We denote a decreasing sequence of control-point counts as $[k_s,\dots,k_0]$ (from finer to coarser). The linear transport operator from coarse level $k_{s-1}$ to fine level $k_s$ is $\mathcal{T}_{s}= \mathcal{O}_{T',k_s}{\mathcal{B}}_{T',k_{s-1}}\in\mathbb{R}^{k_s \times k_{s-1}}$. A fine-level control points admits the decomposition interpolated from coarse level: $\mathcal{P}_{n,k_s}
= \mathcal{T}_s\mathcal{P}_{n,k_{s-1}}+\mathcal{R}_{s}$, where the high-frequency residual is obtained: $
\mathcal{R}_{s}= \big(I_{k_s} - \mathcal{T}_{s}\mathcal{T}_{s}^{+}\big)\mathcal{P}_{n,k_s}$, via the Moore–Penrose pseudo-inverse $\mathcal{T}_{s}^{+}$. The residual from $\mathcal{P}_{n,k_{s-1}}$ to $\mathcal{P}_{n,k_{s-2}}$ is
$
\mathcal{R}_{s-1}
= (I_{k_{s-1}} - \mathcal{T}_{s-1}\mathcal{T}_{s-1}^{+})\mathcal{P}_{n,k_{s-1}}
= \big[(I_{k_{s-1}} - \mathcal{T}_{s-1}\mathcal{T}_{s-1}^{+})\mathcal{T}_{s}^{+}\big]\mathcal{P}_{n,k_s}.$
Similarly, this proceeds recursively until $\mathcal{R}_{1}$.
The coarsest control points are derived by chained projection: $
\mathcal{P}_{n,k_0} = \Big(\prod^{s}_{i=1} {{\mathcal{T}_{i}^{+}}}\Big)\mathcal{P}_{n,k_s}$.  Collecting all multilevel residuals and the coarsest coefficients forms the stacked embedding:
\begin{equation}
\mathcal{E}_{k} = [\mathcal{R}_{s} ,\dots,\,\mathcal{R}_{1},\,\mathcal{P}_{n,k_0}]
= \mathcal{W}_{k}{\mathcal{P}}_{n,k_s}, 
\label{eq: k_0 embedding}
\end{equation}
with the basis $\mathcal{W}_{k}\!\in\!\mathbb{R}^{(\sum_{i=0}^{s} k_i)\times k_s}$ registered as a buffer, $\mathcal{E}_{k}$ is efficiently derived during training via a single matrix multiplication from $\mathcal{P}_{n,k_s}$ padded by $\mathcal{P}_{n,k}$ (see Suppl.). To align with the feature dimension of $\mathbf{F}_0$, the $\mathcal{E}_{k}$ is passed through an MLP to obtain the motion embedding $\mathbf{F}_{k}\!\in\!\mathbb{R}^{n\times c}$.

To compress the embeddings into fewer tokens $n'<n$, and following prior observations \cite{wu2025animateanymesh, zhang2024tapmo, he2025category} that geometrically similar regions of the initial shape tend to exhibit similar motion patterns, we apply farthest point sampling (FPS)~\cite{qi2017pointnet++} to the normal-fused features $\mathbf{F}_0$, producing a reduced set $\mathbf{F}_0' \in \mathbb{R}^{n' \times c}$. The same sampling indices are then applied to each $\mathbf{F}_k$ to obtain $\mathbf{F}_k' \in \mathbb{R}^{n' \times c}$. To capture long-range geometric correspondences, we adopt a cross-attention module that uses $\mathbf{F}_0'$ as the query, the full $\mathbf{F}_0$ as the key, and either $\mathbf{F}_0$ or $\mathbf{F}_{k}$ as the value:
\begin{equation}
\begin{aligned}
\mathbf{F}_0' &= \mathrm{CrossAttn}(\mathbf{F}_0', \mathbf{F}_0, \mathbf{F}_0) + \mathbf{F}_0', \\
\mathbf{F}_{k}' &= \mathrm{CrossAttn}(\mathbf{F}_0', \mathbf{F}_0, \mathbf{F}_{k}) + \mathbf{F}_{k}'.
\end{aligned}
\label{eq:cross_attn}
\end{equation}

By stacking $S$  such cross-attention layers,  we alternately integrate global context into the sparse features $\mathbf{F}_0'$ and $\mathbf{F}_{k}'$, completing spatial compression for both the initial shape and motion.  The compressed $\mathbf{F}_0'$ is projected by a single-layer MLP into the initial latent code $\mathbf{z}_0$ for conditional generation, while motion $\mathbf{F}_{k}'$ is passed through two MLPs to predict the mean $\boldsymbol{\mu}_{k}$ and standard deviation $\boldsymbol{\sigma}_{k}$ respectively. The compact motion latent is then sampled as
$
\mathbf{z}_{k} = \boldsymbol{\mu}_{k} + \boldsymbol{\sigma}_{k} \cdot \epsilon, \epsilon \sim \mathcal{G}(0,1),
$
and $(\boldsymbol{\mu}_{k},\boldsymbol{\sigma}_{k})$ is regularized by the KL divergence $\mathcal{L}_{\mathrm{KL}}$~\cite{Kingma2014} to facilitate subsequent generation training.  For the VAE decoder, $(\mathbf{z}_0,\mathbf{z}_{k})$ are each reprojected by single-layer MLPs and processed by $S$ alternating self-attention layers (cf. \cref{eq:cross_attn}) to recover FPS-sampled initial $\hat{\mathbf{F}}_0'$ and motion $\hat{\mathbf{F}}_{k}'$. To reconstruct the complete motion, we apply cross-attention with full initial feature $\mathbf{F}_0$ as query, sampled initial $\hat{\mathbf{F}}_0'$ as key, and sampled motion $\hat{\mathbf{F}}_{k}'$ as value:
\begin{equation}
\hat{\mathbf{F}}_{k} = \mathrm{CrossAttn}(\mathbf{F}_0,\hat{\mathbf{F}}_0',\hat{\mathbf{F}}_{k}'),
\label{eq:decoder_crossattn}
\end{equation}
which then reconstructs the predicted control points $\hat{\mathcal{P}}_{n,k}$ via an MLP. For stable VAE training, we adopt the Charbonnier loss \cite{barron2019general} to fit the input control points $\mathcal{P}_{n,k}$  as:

\begin{equation}
\mathcal{L}_{\mathrm{Fit}} 
= \mathbb{E}\!\left[\sqrt{\|\hat{\mathcal{P}}_{n,k} - \mathcal{P}_{n,k}\|_F^2 + \delta^2}\right],
\end{equation}
with $\delta = 1\times10^{-3}$ for numerical stability and smooth gradients. Besides, we introduce a correspondence loss to fit the GT sampled differences trajectory 
$\mathcal{V}_{n,T'}$ via B-spline reprojection from predicted control points 
$\hat{\mathcal{V}}_{n,T'} = \mathcal{B}_{T',k} \hat{\mathcal{P}}_{n,k}$:
\begin{equation}
\mathcal{L}_{\mathrm{Corr}} 
= \mathbb{E}\!\left[\sqrt{\|\hat{\mathcal{V}}_{n,T'} - \mathcal{V}_{n,T'}\|_F^2 + \delta^2}\right],
\end{equation}
with $T'=k=16$ during training  for comparability with \cite{wu2025animateanymesh}. Empirically, $\mathcal{L}_{\mathrm{Corr}}$ decreases faster than $\mathcal{L}_{\mathrm{Fit}}$, providing effective guidance during early training.  We also introduce a local rigidity loss, inspired by skeletal representations, to enforce consistent local lengths across frames:
\begin{equation}
\mathcal{L}_{\mathrm{Rigid}} =
\mathbb{E}
\Bigg[
\sqrt{
\Big(
r_t(i,j) - r_{t-1}(i,j)
\Big)^2 + \delta^2
}
\Bigg],
\label{Eq:Rigid}
\end{equation}
where $r_t(i,j)$ denotes the $L_2$ motion distance from point $i$ to its KNN neighbor $j$ at time $t$, computed from $\hat{\mathcal{V}}_{n,T'}$. 
The total VAE training objective for parameters $\theta_{\mathrm{vae}}$ is then:
\begin{equation}
    \mathcal{L}_\mathrm{VAE}(\theta_{\mathrm{vae}}) = \mathcal{L}_\mathrm{Fit} + \lambda_{1}\mathcal{L}_{\mathrm{Corr}} +\lambda_2 \mathcal{L}_{\mathrm{Rigid}}+\lambda_3 \mathcal{L}_{\mathrm{KL}},
\end{equation}
with weights $\lambda_1=3\times10^{-1}$, $\lambda_2=10^{-1}$ and $\lambda_3=2\times10^{-5}$.

\subsection{Shape-Driven Text-to-Motion Generation}
\label{sec: generation_model}
After the VAE encoding, we build a generative model conditioned (\cref{fig:overview}) on the initial shape and text to infer the motion latent distribution. To ensure stability, following \cite{wu2025animateanymesh,zhang2025gaussian}, we normalize the initial latent $\mathbf{z}_0$ and motion latent $\mathbf{z}_k$ to $\tilde{\mathbf{z}}_0$ and $\tilde{\mathbf{z}}_k$ using their running means $(\tilde{\mathbf{\mu}}_0, \tilde{\mathbf{\mu}}_k)$ and standard deviations $(\tilde{\mathbf{\sigma}}_0, \tilde{\mathbf{\sigma}}_k)$ computed from the training set.

We employ rectified flow-matching (FM) transport, integrated from learnable velocity network $v_{\theta_\mathrm{vel}}$ \cite{liuflow}, to sample the normalized motion latent \(\tilde{\mathbf{z}}_{k}\) at \(\tau=1\) from standard Gaussian noise \(\epsilon\) at \(\tau=0\). The trajectory is defined as $
\tilde{\mathbf{z}}_{k}^{\tau} = \tau\,\tilde{\mathbf{z}}_{k} + (1-\tau)\,\epsilon, \tau\in[0,1]$. With the initial $\tilde{\mathbf{z}}_0$ and textual prompt $y$ as condition, the conditional velocity $v_{\theta_{\mathrm{vel}}}(\tilde{\mathbf{z}}_{k}^{\tau}|\tau,\tilde{\mathbf{z}}_0,y)$ is trained with the following loss:
\begin{equation}
    \mathcal{L}_{\mathrm{FM}}(\theta_{\mathrm{vel}})= \mathbb{E}||v_{\theta_{\mathrm{vel}}}(\tilde{\mathbf{z}}_{k}^{\tau}|\tau,\tilde{\mathbf{z}}_0,y) - \mathbf{u}||_F^2,
\end{equation}
where the target velocity is $\mathbf{u} = \tilde{\mathbf{z}}_{k} - \epsilon$, and the sampling time $\tau$ follows a cosine schedule \cite{esser2024scaling}. Following \cite{wu2025animateanymesh}, the velocity network $v_{\theta_{\mathrm{vel}}}$ is based on Diffusion Transformers (DiTs)~\cite{peebles2023scalable, esser2024scaling} with $S'=12$  DiT blocks. To enhance spatial–motion correspondence, following Wan~\cite{wan2025wan}, $\tilde{\mathbf{z}}_0$ is concatenated with $\tilde{\mathbf{z}}_{k}^{(\tau)}$ as input and used together with text embeddings from the CLIP text encoder~\cite{radford2021learning} as conditions via decoupled cross attention \cite{ye2023ip}. See Suppl. for details.

\subsection{Sequence Generation at Inference}

With the model parameters $\theta=(\theta_{\mathrm{vae}},\theta_{\mathrm{vel}})$ trained, feed-forward inference can be performed. Given an initial mesh and text prompt, the initial latent $\mathbf{z}_0$ is computed via the VAE encoder with vertices and normals. A clean latent $\tilde{\mathbf{z}}_{k}$ is then sampled from noise $\epsilon$ via the flow-based ODE~\cite{torchdiffeq}, applying classifier-free guidance with strength $\gamma=3.0$, conditioned on the normalized initial latent $\tilde{\mathbf{z}}_0$ and text $y$. The original $\mathbf{z}_{k}$ is restored using preserved statistics $(\tilde{\mathbf{\mu}}_{k},\tilde{\mathbf{\sigma}}_{k})$. The VAE decoder then predicts control points $\hat{\mathcal{P}}_k$ via the normal-fused vertex feature $\mathbf{F}_0$ in \cref{eq:decoder_crossattn}. The generated vertex sequence is then obtained via B-spline reprojection:
\begin{equation}
\label{eq: evaluation}
\hat{\mathbf{V}}_{1:T} = \mathbf{V}_0 + \mathcal{B}_{T,k}\hat{\mathcal{P}}_{k},
\end{equation}
where the basis $\mathcal{B}_{T,k}$ can be built for any $T$, enabling mesh sequences of arbitrary length for broader applications.

%% file: sec/4_experiment.tex
\section{Experiments}
\label{sec:experiment}

\begin{figure*}[!htb]
    \centering
\includegraphics[width=1\linewidth]{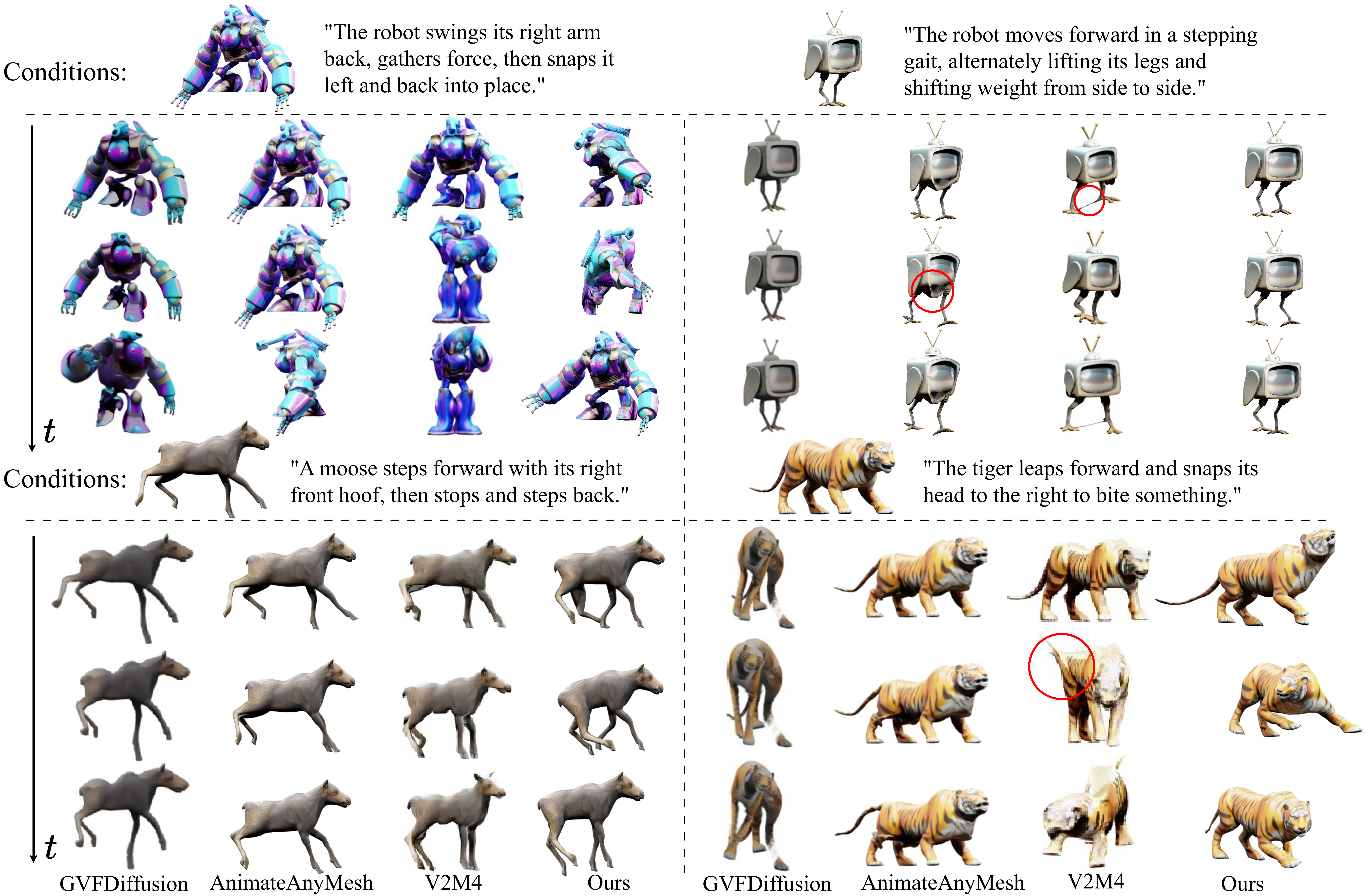}
\vspace{-7mm}
\caption{\textbf{Qualitative Comparisons.} Our method (\methodname) results in superior motion quality and is more aligned with the user-provided text prompts. 
Artifacts for the baseline methods are highlighted in red. 
Please see the supplementary material for additional results.
}
\vspace{-5mm}
\label{fig:qualitative_comparison}
\end{figure*}

\subsection{Dataset Preparation}
To validate our method, we curated a new dataset called \textbf{BIMO}, comprising 38,944 motion sequences with a total of 3,682,790 frames. It integrates diverse mesh motion sequences from DeformingThings4D Animals~\cite{li20214dcomplete}, ObjaverseV1~\cite{deitke2023objaverse}, and ObjaverseXL~\cite{deitke2023objaverseXL}. Asset IDs from ObjaverseV1 were curated by Diffusion4D~\cite{liang2024diffusion4d}, while meta IDs from ObjaverseXL were filtered by GVFDiffusion~\cite{zhang2025gaussian}. We reserve 400 sequences for design validation. Each variable-length raw sequence from 5 to 200 frames, is converted into 16-frame B-spline control points following \cref{subsec:bspline-compress}. For captions, DeformingThings4D provides human-annotated motion descriptions from OmniMotionGPT~\cite{yang2024omnimotiongpt}, with three captions per motion sequence. For Objaverse, we use an automatic captioning pipeline based on the multimodal VLM GPT-5, which generates three captions per rendered video in the style of DeformingThings4D. The pipeline includes a caption generator and an inspector to ensure correctness and consistency (see Suppl.~for details). We will release BIMO upon acceptance.

\subsection{Implementation Details} 
In \cref{eq: k_0 embedding}, we support the simplest B-spline curves with $k_0=4$. Following \cite{wu2025animateanymesh} during training, we randomly downsample $n = 4096$ points which is further compressed into $n' = 512$ tokens, while the B-spline VAE employs an encoder with $S = 8$ cross-attention layers and a decoder  with $S = 8$ self-attention layers. 
The VAE is optimized with Adam \cite{loshchilovdecoupled} using a learning rate of $1\times10^{-4}$ with weight decay 0.01. 
The velocity field is trained with a learning rate of $1\times10^{-5}$, linearly warmed up from 0 over the first 1000 iterations. 
During generative training, following \cite{wu2025animateanymesh}, we use a lightweight pre-trained CLIP ViT-L/14 \cite{radford2021learning} with a maximum text length of 77 for text embeddings, and we set a 0.1 dropout rate for both conditioning inputs. 
Both models apply random-noise and rotation augmentations to initial points, normals, points differences, and control points in each iteration. 
We use FP16 mixed precision and FlashAttention \cite{dao2023flashattention2} to accelerate training. 
Both the VAE (\cref{subsec:bsplinevae}) and the velocity network (\cref{sec: generation_model}) are trained on eight NVIDIA L40 GPUs for seven days.

\definecolor{LightCyan}{rgb}{0.88,1,1}
\begin{table}[!t]
\centering
\caption{\textbf{Quantitative Comparisons.} Our approach not only surpasses existing methods across multiple metrics (\cref{exp:metrics}), but also requires less time and peak GPU memory (PG) on an A100 GPU.
}
\label{tab:Quantitative_Comparisons}
\vspace{-3mm}
\resizebox{\linewidth}{!}{
\begin{tabular}{l|ccccc}
\toprule
\multirow{2}{*}{Methods} & \multicolumn{5}{c}{\textbf{Vbench}}\\
\cline{2-6}
& OC $\uparrow$ & SC $\uparrow$ & TF $\uparrow$ & AQ $\uparrow$ & $DD \uparrow$\\
\midrule
GVFDiffusion \cite{zhang2025gaussian} & 0.167 & 0.920 & 0.986 & 0.505 & 0.650 \\
AnimateAnyMesh \cite{wu2025animateanymesh} & 0.155 & \textbf{0.951} & 0.993 & 0.514 & 0.100 \\
V2M4 \cite{chen2025v2m4} & 0.175 & 0.876 & 0.986 & 0.478 & 0.750 \\
\rowcolor{LightCyan}
\textbf{\methodname} (Ours) & \textbf{0.187} & 0.948 & \textbf{0.995} & \textbf{0.529} & \textbf{0.800} \\
\midrule
\multirow{2}{*}{Methods} & \multicolumn{3}{c|}{\textbf{User Study}} &\multicolumn{2}{c}{\textbf{Efficiency}} \\
\cline{2-6}
& TA $\uparrow$ & MP $\uparrow$ & \multicolumn{1}{c|}{ME $\uparrow$} & Time $\downarrow$ & PG (GB) $\downarrow$ \\
\midrule
GVFDiffusion \cite{zhang2025gaussian} & 2.343 & 2.300 & \multicolumn{1}{c|}{2.443} & 2.141m & 14.057 \\
AnimateAnyMesh \cite{wu2025animateanymesh} & 2.314 & 2.686 & \multicolumn{1}{c|}{2.443} & 16.847s & 3.102 \\
V2M4 \cite{chen2025v2m4} & 2.876 & 2.714 & \multicolumn{1}{c|}{3.048} & 1.672h & 48.416 \\
\rowcolor{LightCyan}
\textbf{\methodname} (Ours) & \textbf{4.095} & \textbf{4.062} & \multicolumn{1}{c|}{\textbf{4.048}} & \textbf{4.383s} & \textbf{1.246} \\
\bottomrule
\end{tabular}
}
\vspace{-5mm}
\end{table}

\subsection{Comparisons}
\textbf{Baselines.} We compare against three SOTAs: (1) \textit{AnimateAnyMesh}~\cite{wu2025animateanymesh}, the most related work, which performs feed-forward generation from an initial mesh user-provided text prompt; (2) \textit{GVFDiffusion}~\cite{zhang2025gaussian}, which generates dynamic 3D Gaussians from videos; and (3) \textit{V2M4}~\cite{chen2025v2m4}, which reconstructs dynamic meshes from monocular videos. To mitigate artifacts introduced during video preparation, we employ the high-quality video generator Kling~\cite{kling2024} as the video source for the latter two methods.

\vspace{1mm}
\label{exp:metrics}
\noindent\textbf{Metrics.} We evaluate on 20 randomly selected static meshes (twice the number used in AnimateAnyMesh), including AI-generated ones from Meshy~\cite{meshy2025discover} and existing assets without ground-truth motion. Following \cite{wu2025animateanymesh, jiang2024animate3d}, we render 16-frame animations from fixed viewpoints for fair comparison and assess perceptual metrics using VBench~\cite{huang2023vbench}: (1) \textit{overall consistency} (OC) for prompt-video alignment, (2) \textit{subject consistency} (SC) for appearance coherence, (3) \textit{temporal flickering} (TF) for temporal stability, (4) \textit{aesthetic quality} (AQ) for shape fidelity, and (5) \textit{dynamic degree} (DD) for motion plausibility. We also report average generation time and peak GPU memory (PG) per mesh animation. To further evaluate motion quality, we conduct a \textbf{user study} with 20 participants. Each participant reviews 10 cases and rates each animation on a 5-point Likert scale (1 = Very Poor, 5 = Excellent) for three aspects: (1) \textit{text-to-motion agreement} (TA), (2) \textit{motion plausibility} (MP), and (3) \textit{motion expressiveness} (ME). Final scores are averaged across all ratings (see Suppl.~for details).

\vspace{1mm}
\noindent\textbf{Results.} In \cref{fig:qualitative_comparison}, SOTAs show clear limitations: the feed-forward GVFDiffusion often produces incorrect shapes or textures (\eg Tiger) and fails to reproduce desired motion patterns; AnimateAnyMesh yields only slight motion due to its small fixed-frame cropping during training, which cause higher subject consistency (which has inflated scores for near static video) but very low dynamic degree in \cref{tab:Quantitative_Comparisons}. The optimization-based V2M4 is sensitive to video content and struggles to maintain object consistency across frames (\eg Robot and Tiger).  Our method produces more accurate motions (\eg Moose) that better match text descriptions and contain fewer artifacts,  thanks to efficient motion representation based on B-splines and high-quality annotations during training. In \cref{tab:Quantitative_Comparisons}, our approach outperforms in most Vbench metrics and achieves the highest scores across all user-study metrics, demonstrating better prompt–motion adherence and more expressive, plausibly coherent motion. Moreover, our method is more stable and efficient: as mesh vertices increase from 9\text{K} to 24\text{K}, AnimateAnyMesh’s generation time and peak GPU memory rise sharply from $6.4$ s to $20.0$ s and $1.6$ GB to $3.6$ GB due to heavy topology processing, whereas ours only slightly increase from $3.7$ s to $4.6$ s and $1.1$ GB to $1.3$ GB.

\subsection{Ablations}
\begin{figure}[!htb]
    \centering
    \includegraphics[width=1\linewidth]{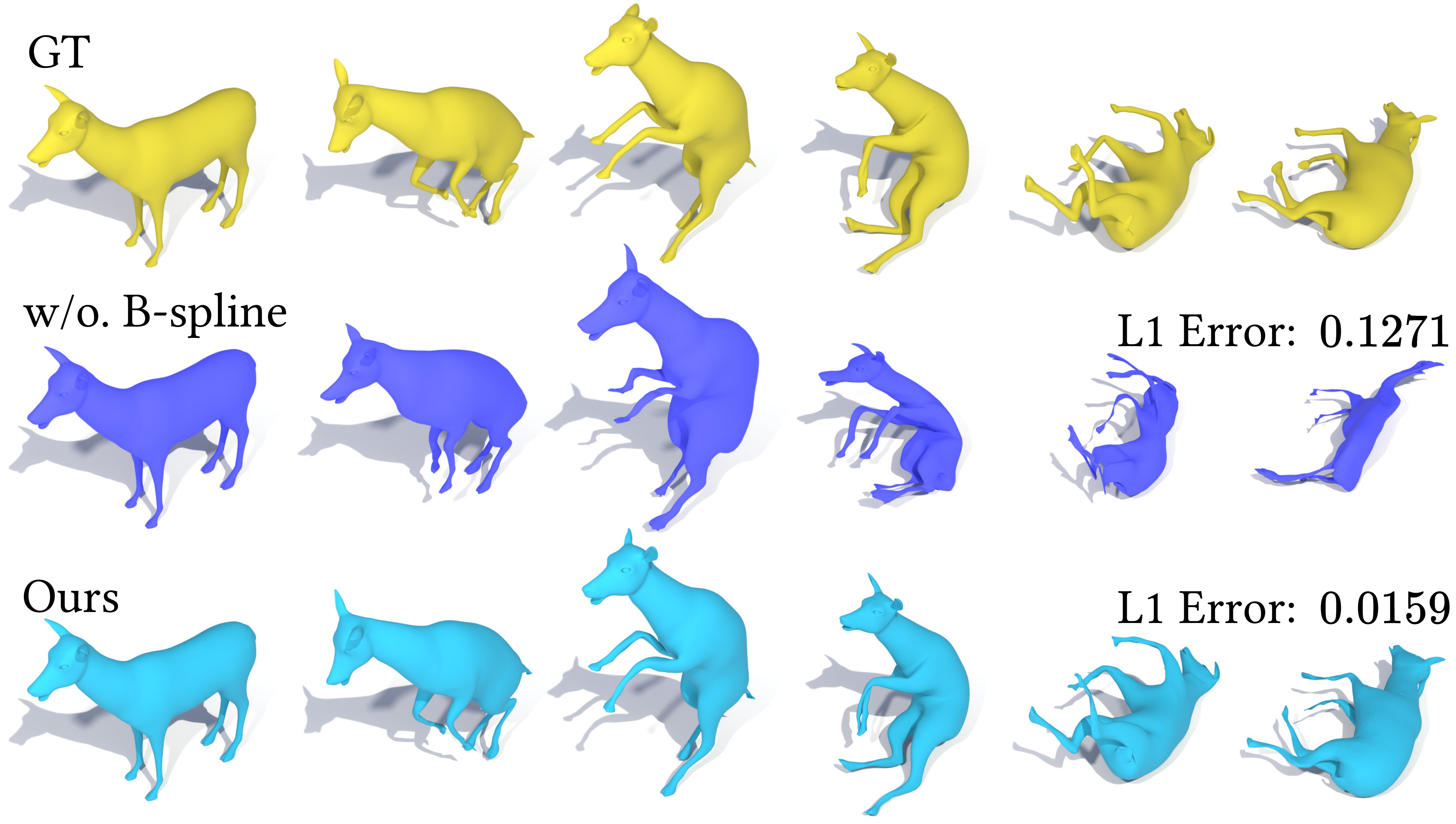}
\vspace{-8mm}
\caption{\textbf{B-spline Ablation.} B-spline interpolation from predicted control points achieves lower L1 error over the entire sequence than linear interpolation on sampled raw differences.}
\vspace{-3mm}
\label{fig:bspline_ablation}
\end{figure}

\begin{figure}[!htb]
    \centering
    \includegraphics[width=1\linewidth]{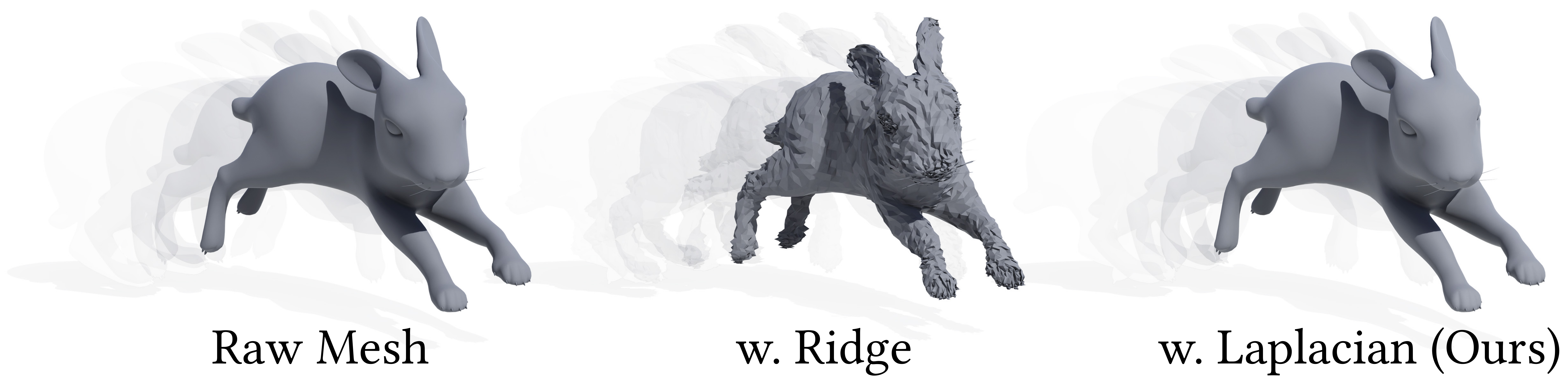}
    \vspace{-7mm}  \caption{\textbf{Laplacian Regularizer $\mathbf{L}$.} B-spline $\times2$ interpolation comparison from a $T=10$ mesh sequence comparing Ridge \cite{hoerl1970ridge} and our Laplacian regularizer $\mathbf{L}$. $\mathbf{L}$ produces more natural results.} 
\label{fig:ablation_lambda}
%\vspace{-3mm}
\end{figure}

\noindent \textbf{B-spline Representation for Variable-length Sequences.} We encode variable-length motion sequences using B-splines and evaluate on the BIMO validation set (400 sequences of varying lengths). 
For fair comparison, the baselines uniformly downsample each sequence to 16 frames, the same number as our control points with the same VAE architecture, and linearly interpolate them back to the original length. 
In \cref{tab:abl}, all B-spline VAE variants achieve higher reconstruction fidelity than the linear interpolation baselines (first two rows). 
In \cref{fig:bspline_ablation}, under the same 16-fixed VAE constraint, our method’s interpolations via \cref{eq: evaluation} more faithfully reproduce the GT sequence, whereas the baseline's motion is inconsistent between the discrete samples, resulting in reconstruction failures. 
We further evaluate the proposed Laplacian regularization in \cref{fig:ablation_lambda}, which produces more natural interpolations under fewer frames ($T<k$) compared to other regularization types, \eg Ridge. % \cite{mcdonald2009ridge,marquardt1975ridge}.

\begin{table}[!t]
  \centering
  \caption{\textbf{VAE Quantitative Ablations.} “Rec Error” indicates the mean L1 error ($\times 10^{-2}$) averaged per trajectory and per instance.}
  \vspace{-5pt}
  \resizebox{0.9\linewidth}{!}{
  \begin{tabular}{ccccc|c}
    \toprule
    B-Spline & NF & Control-PE & $\mathcal{L}_\mathrm{Corr}$ & $\mathcal{L}_\mathrm{Rigid}$ & Rec Error $\downarrow$ \\
    \midrule
    \xmark & \xmark& \xmark&\xmark&\xmark&3.237\\
    \xmark&\cmark&\xmark &\cmark &\cmark&2.674\\
    \cmark & \xmark&\cmark&\cmark&\cmark&1.328\\
    \cmark &\cmark&\xmark&\cmark&\cmark&1.648\\
    \cmark &\cmark&\cmark &\xmark&\cmark&1.303\\
    \cmark &\cmark&\cmark &\cmark&\xmark&1.349\\
    \rowcolor{LightCyan} \textbf{\cmark} &\textbf{\cmark}&\textbf{\cmark} &\textbf{\cmark}& \textbf{\cmark}&\textbf{1.078}\\
    \bottomrule
  \end{tabular}}
  \label{tab:abl}
  \vspace{-10pt}
\end{table}

%\vspace{1mm}

\begin{figure}[!htb]
    \centering
\includegraphics[width=1\linewidth]{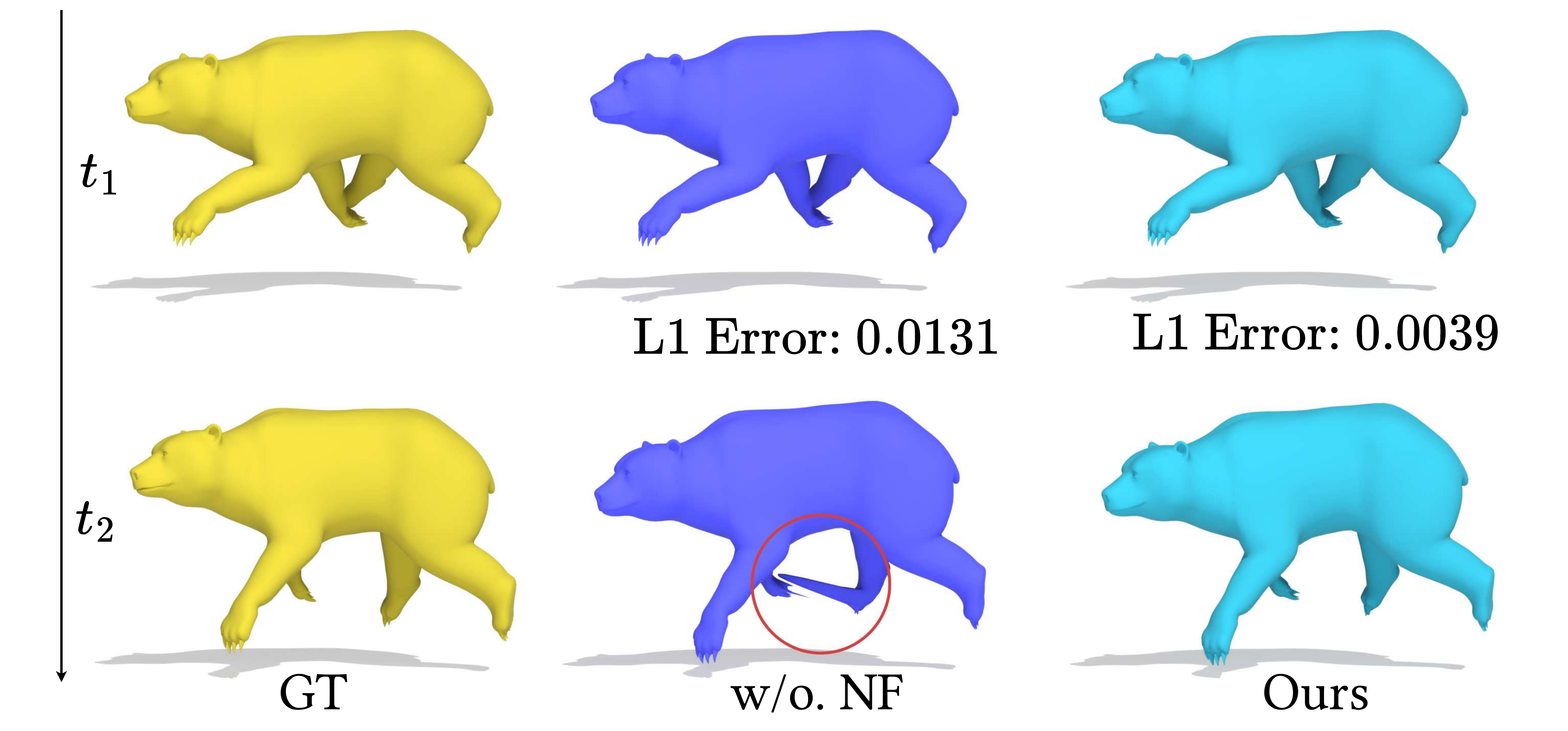}
\vspace{-7mm} 
\caption{\textbf{Normal Fusion (NF) Ablation.} NF effectively separates nearby distinct parts. 
Without it, artifacts appear (see red).}
\vspace{-5mm} 
\label{fig:normal_ablation}
\end{figure}

\vspace{1mm}
\noindent \textbf{Normal Fusion.} Our normal fusion strategy (NF in \cref{tab:abl}) efficiently leverages local geometric cues from normals to distinguish spatially close but different motion units (\cref{fig:normal_ablation}). See the Suppl.~for additional ablations.

\begin{figure}[!htb]
    \centering
    
    \includegraphics[width=1\linewidth]{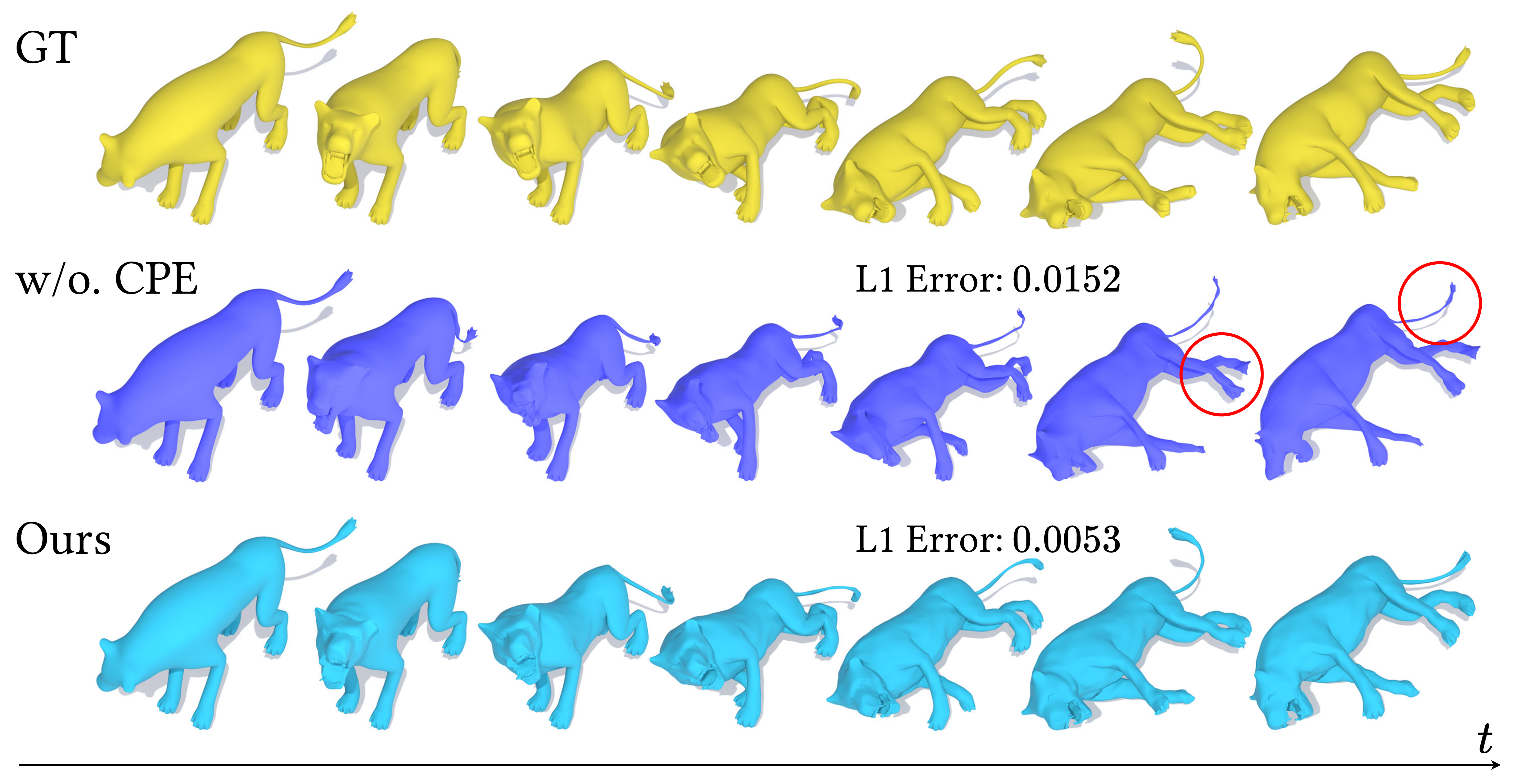}
    \vspace{-7mm} \caption{\textbf{Control-PE Ablation.} Our control-PE (Bottom) captures fine motion (\eg the lion’s tail) and reduces artifacts (red circles) compared to conventional position encoding  (Middle).}
    \vspace{-6mm}
    \label{fig:CPE_ablation}
\end{figure}

\vspace{1mm}
\noindent \textbf{Control Point Embedding.} Our multi-level control-point embedding (Control-PE in \cref{tab:abl}) preserves both coarse and fine motion details better than the conventional frequency-based position encoding used in AnimateAnyMesh~\cite{wu2025animateanymesh}. As shown in \cref{fig:CPE_ablation}, Control-PE (Bottom) more accurately reproduces ground-truth motion (\eg the lion’s tail).

\begin{figure}[!t]
    \centering
    \includegraphics[width=1\linewidth]{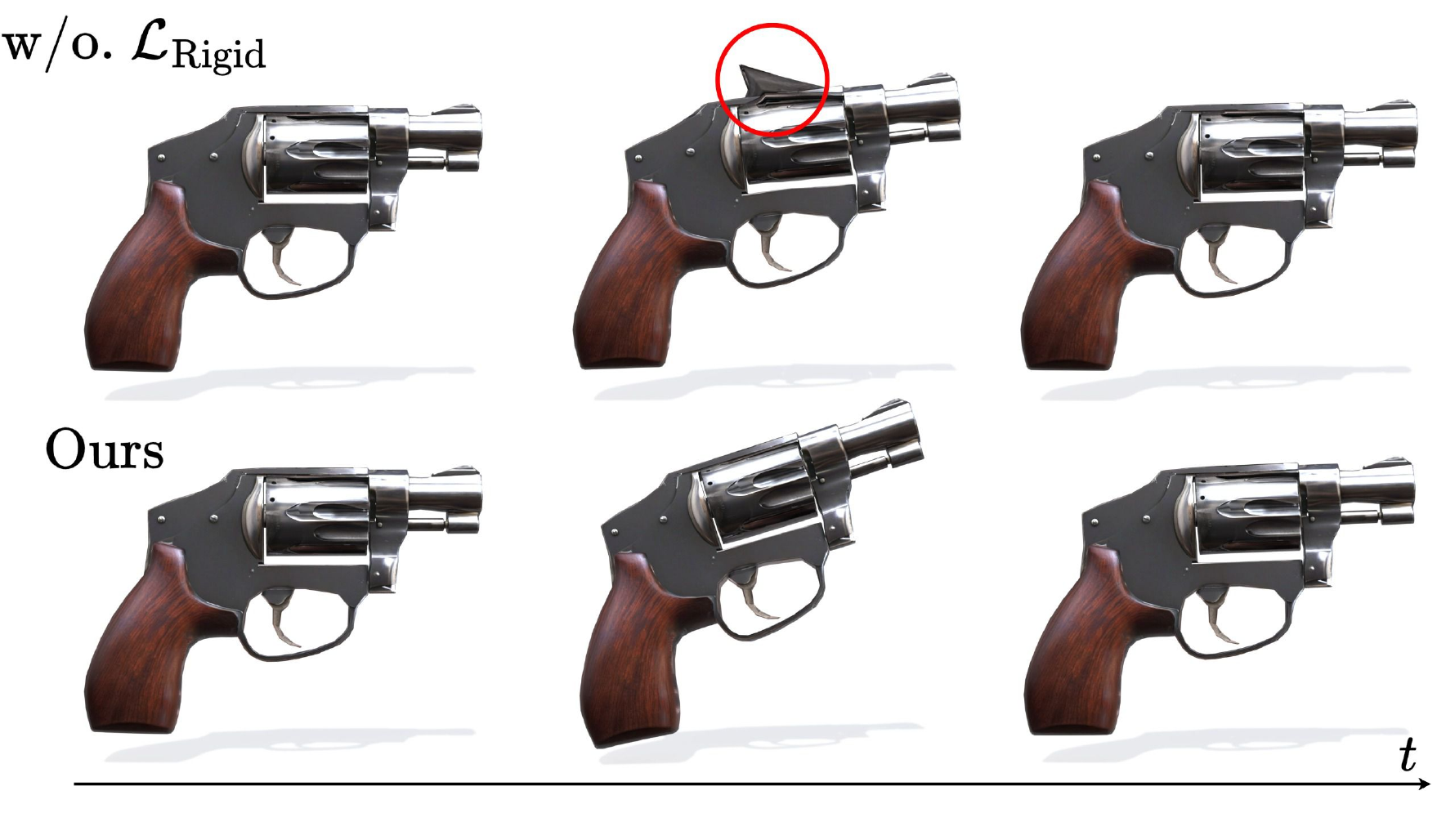}
\vspace{-7mm}
\caption{\textbf{$\mathcal{L}_{\mathrm{Rigid}}$ Ablation.} Our local rigid loss reduces the red-circled artifacts during motion generation. Prompt: \textit{``The pistol tilts up to an angled pose, then returns to forward-facing."}}
\vspace{-6mm}
\label{fig:rigid_ablation}
\end{figure}

\vspace{1mm}
\noindent \textbf{Local Rigidity.} We target mesh motion generation for 3D character animation,  which inherently requires preserving local rigidity. The proposed local rigid loss $\mathcal{L}_\mathrm{Rigid}$ (\cref{Eq:Rigid}) enforces piecewise consistency with the initial geometry across frames, helping retain shape identity during motion generation, as shown in \cref{fig:rigid_ablation}.

\begin{figure}[!htb]
    \centering
\vspace{-2mm}  \includegraphics[width=1\linewidth]{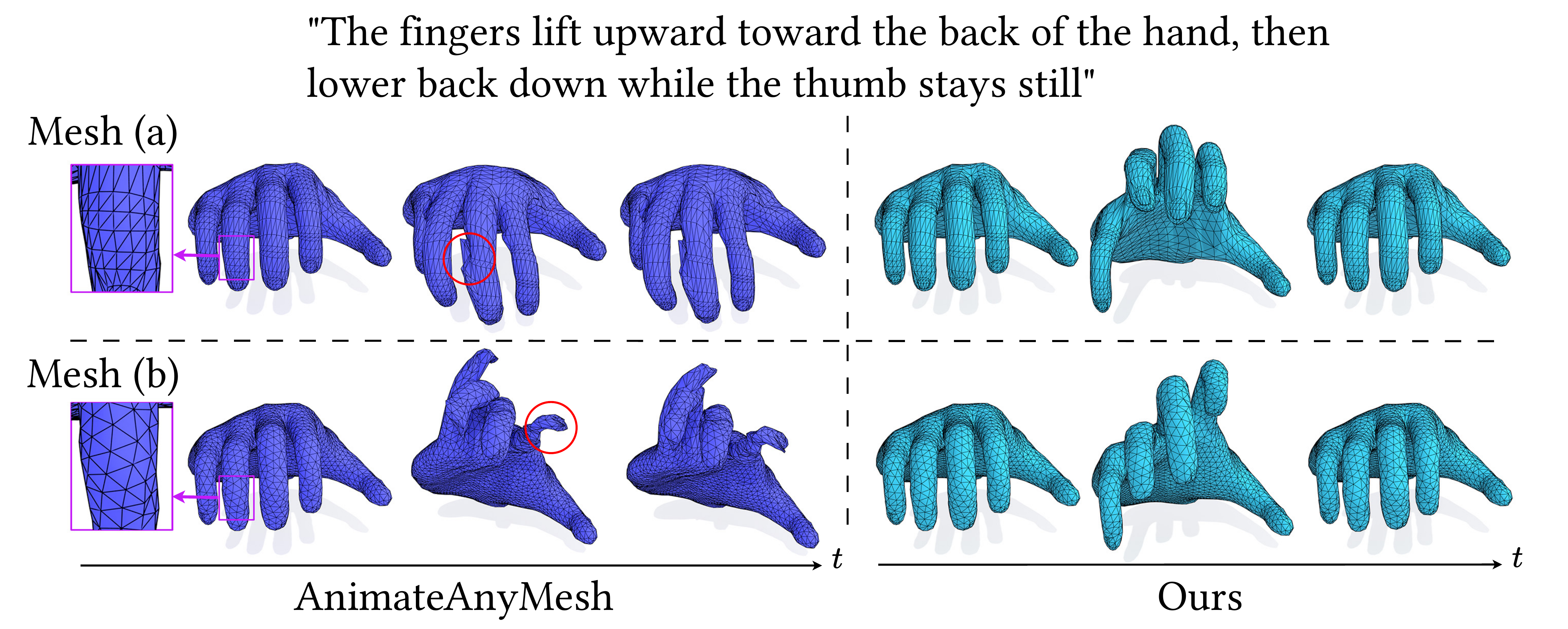}
\vspace{-7mm}
\caption{\textbf{Robustness to Meshing.} Ours produces consistent motion under mesh changes (Top to Bottom), whereas AnimateAnyMesh exhibits unstable motion and artifacts (red circles).}
\label{fig:topology_robust}
\vspace{-5mm}
\end{figure}

\noindent \textbf{Robustness to Meshing.} Our dense-point training and normal fusion strategies are independent of fixed mesh topology, making our method robust to mesh changes (see \cref{fig:topology_robust}) and capable of generating motion consistent with the user-provided input text, while AnimateAnyMesh fails.

%\vspace{-10pt}

%% file: sec/6_conclusion.tex
\section{Discussion}
\label{sec:conclusion}
Like all other generative tasks, improvements in motion generation depend on the availability of large and  diverse dynamic sequences and sufficient compute. 
\methodname nonetheless offers a complementary, staged contribution to character motion generation by demonstrating how to obtain continuous, compact, and efficient motion representations from variable-length sequences that can be used with constrained fixed-capacity models. 
This representation-centric view yields more expressive and plausible motions than SOTAs, and we expect these advantages to persist even as more powerful universal motion-generation models emerge. 
Extensive experiments and ablations verify the effectiveness of the proposed designs.

\noindent \textbf{Limitations.} 
\methodname may fail to represent high-frequency details for very complex motions unless more control points are used. It also does not support topology-changing motion due to our fixed-mesh assumption.

%% file: suppl.tex
% \maketitle
\maketitlesupplementary

\appendix
\setcounter{table}{0}
\renewcommand{\thetable}{A\arabic{table}}
\setcounter{figure}{0}
\renewcommand{\thefigure}{A\arabic{figure}}

%\section{Organization}
%In this paper, we introduce \methodname, a fast, feed-forward B-spline–based framework for dynamic 3D character generation. By representing motion with B-splines, our method produces continuous, high-quality, and expressive motion trajectories that faithfully follow rich textual prompts, outperforming state-of-the-art approaches. 
This supplementary material provides additional implementation details, further experiments and results, outlines the user study design and provides statistics, additional ablations (\eg extended normal-fusion ablations), and describes our dataset captioning pipeline. %, offering a comprehensive understanding of our approach.
We also encourage readers to view the \textbf{accompanying video} for demonstrations of the dynamic results.

\section{Additional Implementation Details}
\subsection{Velocity Network}
\begin{figure}[!htb]
    \centering
    \includegraphics[width=1\linewidth]{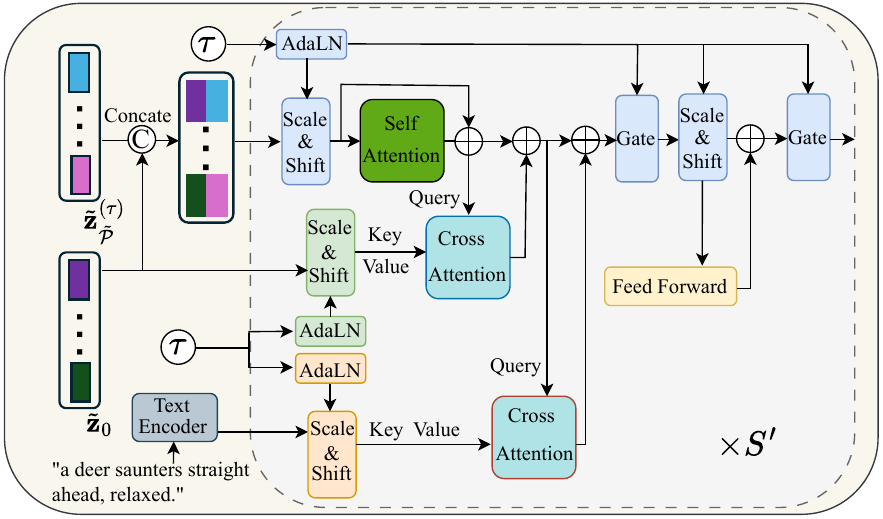}
\caption{\textbf{Velocity Architecture.} The network $v_{\theta_{\mathrm{vel}}}$ comprises $S'$ stacked layers with self- and cross-attention modules.}
\label{fig:velo_Pipeline}
\end{figure}

We implement the velocity field $v_{\theta_{\mathrm{vel}}}$ (described in Sec.~\textcolor{myblue}{3.6} in the paper)  based on Diffusion Transformer (DiTs) \cite{peebles2023scalable} (see \cref{fig:velo_Pipeline}). To promote spatial–motion correspondence, we follow the open-source image-to-video model Wan~\cite{wan2025wan} to concatenate normalized latents along the channel dimension as $\tilde{\mathbf{z}}_{\parallel}=[\tilde{\mathbf{z}}_0,\tilde{\mathbf{z}}_{\tilde{\mathcal{P}}}^{(\tau)}]$. Text prompts are encoded using a pretrained CLIP text encoder \cite{rombach2022high} to obtain text embeddings. Following \cite{esser2024scaling}, we apply separate AdaLN \cite{peebles2023scalable} parameters conditioned on the timestep $\tau$ to modulate the concatenated latents, the  latent $\tilde{\mathbf{z}}_0$, and the text embedding independently. The rescaled $\tilde{\mathbf{z}}_{\parallel}$ first passes through self-attention, then through cross-attention blocks that fuse information from the initial shape and the text embedding. This modality-specific AdaLN dynamically modulates each conditioning signal to vary modality-specific influence across sampling timesteps. After $S'=12$ such DIT attention blocks, the network decodes the control-point velocities $v_{\theta_{\mathrm{vel}}}(\tilde{\mathbf{z}}_{\tilde{\mathcal{P}}}^{(\tau)}\mid\tau,\tilde{\mathbf{z}}_0,y)$.

\subsection{Training Details}
To reduce I/O cost during training, 
and to speed up training,  each dense  point set ($N=200\text{K}$) is randomly split into ten files of $20\text{K}$ points stored separately in preprocessing, where all initial shapes are normalized into range $[-0.9, 0.9]$. In our control-point embedding, we adopt a multi-level hierarchical strategy with a decreasing sequence (from finer to coarser) of control-point counts $[k_s,\ldots,k_0]=[17,15,13,11,9,7,5,4]$. When the input control-point sequence has length $k=16$, we pad it to $17$ by replicating the last control point to ensure compatibility.

\section{Additional Experiments}
\subsection{Additional Comparisons}
\begin{figure*}[!htb]
    \centering
\includegraphics[width=1\linewidth]{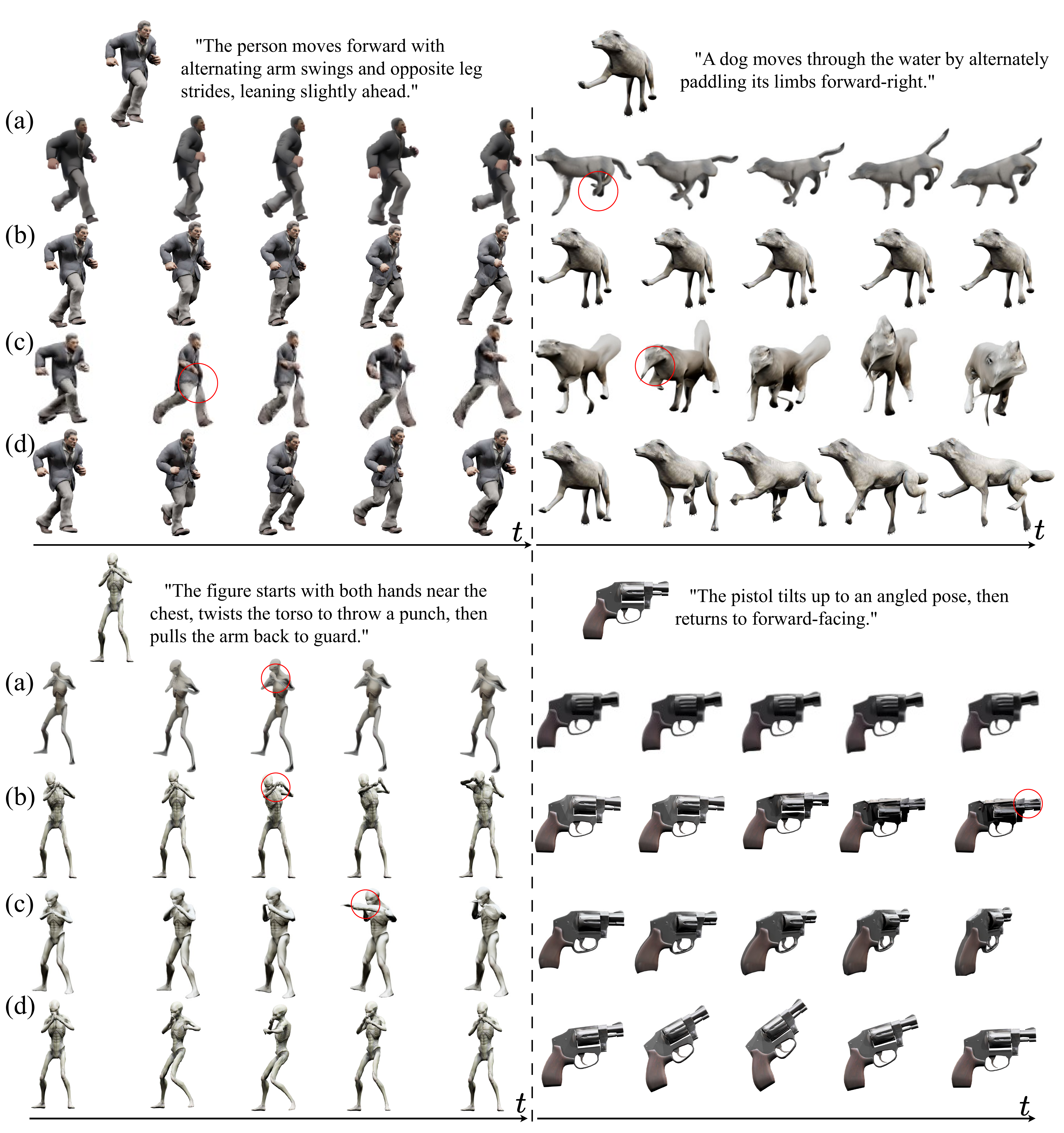}

\caption{\textbf{Additional Qualitative Comparisons.} (a) GVFDiffusion \cite{zhang2025gaussian}, (b) AnimateAnyMesh \cite{wu2025animateanymesh}, (c) V2M4 \cite{chen2025v2m4}, (d) BiMotion (Ours). \methodname yields more text-aligned, higher-quality motion. 
Visual artifacts are illustrated in red.
}
\label{fig:addtional_qualitative_comparison}
\end{figure*}

\begin{figure*}[!htb]
    \centering
\includegraphics[width=1\linewidth]{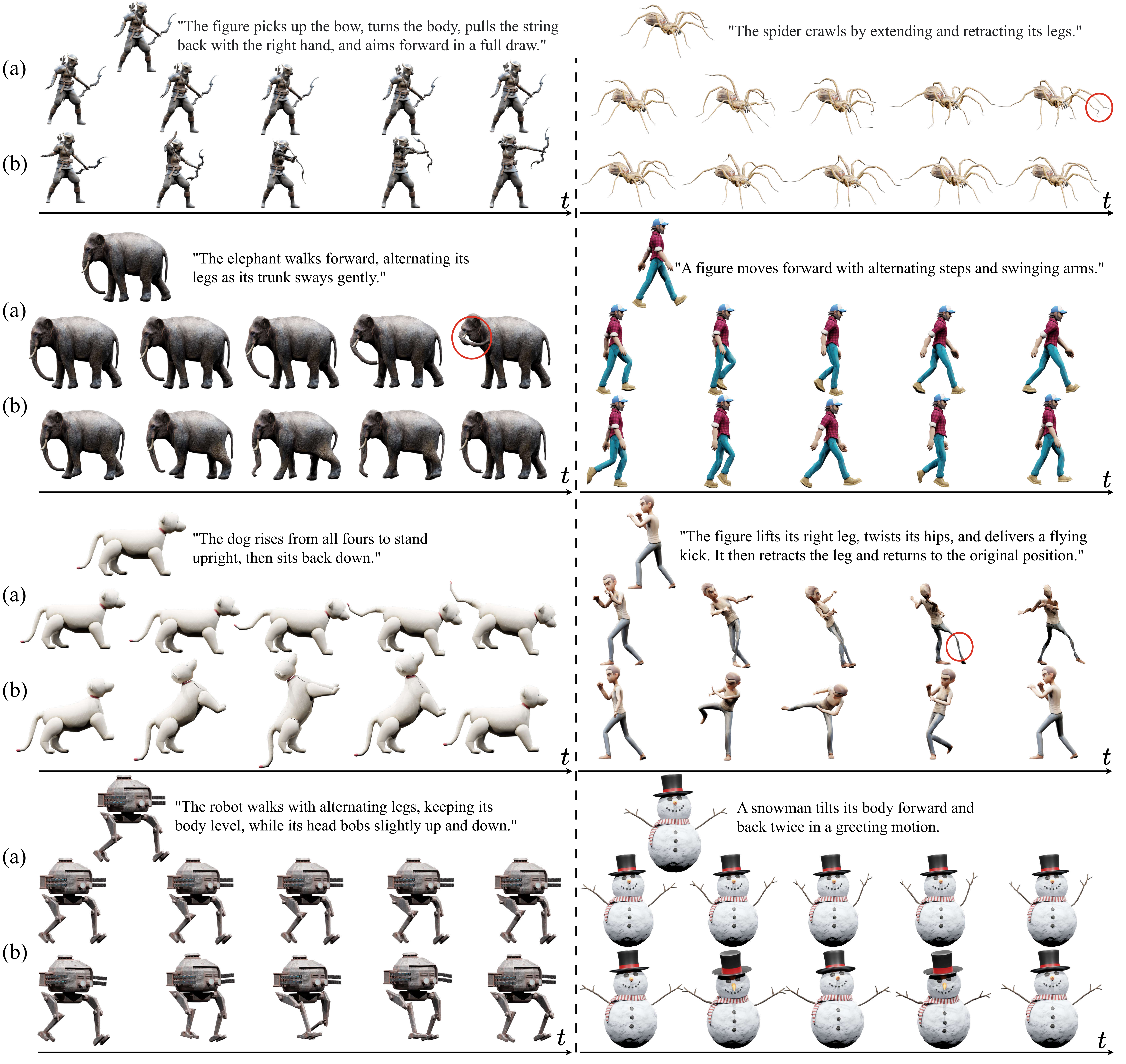}

\caption{\textbf{Additional Qualitative Comparisons with AnimateAnyMesh.}  (a) AnimateAnyMesh \cite{wu2025animateanymesh} and (b) BiMotion (Ours).
}
\label{fig:addtional_qualitative_compariso_w_animate}
\end{figure*}

To provide a comprehensive comparison of mesh animation approaches, we include additional qualitative results against three baseline methods in \cref{fig:addtional_qualitative_comparison}. We further evaluate additional cases against the closest competitor, AnimateAnyMesh~\cite{wu2025animateanymesh}, using multi-mesh sources from Objaverse \cite{deitke2023objaverse}, Sketchfab \cite{sketchfab_free_models}, and AI-generated assets from Meshy \cite{meshy2025discover} and Hunyuan3D \cite{zhao2025hunyuan3d}. Additional comparisons with AnimateAnyMesh are presented here in \cref{fig:addtional_qualitative_compariso_w_animate} and in Fig.~\textcolor{myblue}{1} of the main paper. 
Across all settings, the results consistently support our main observation that method produces richer text-aligned motion, more natural dynamics, and more expressive animations compared to  existing approaches.

For certain walking cases, such as Elephant and Robot in \cref{fig:addtional_qualitative_compariso_w_animate}, the motion may appear to be ``in-place.” This is because many dynamic 3D assets depict walking using in-place cycles by default. 
Artists typically add the global displacement (\ie the translation) only when integrating the character into a specific scene or environment.

\subsection{Additional Baselines}
\begin{figure}[!htb]
    \centering
\includegraphics[width=1\linewidth]{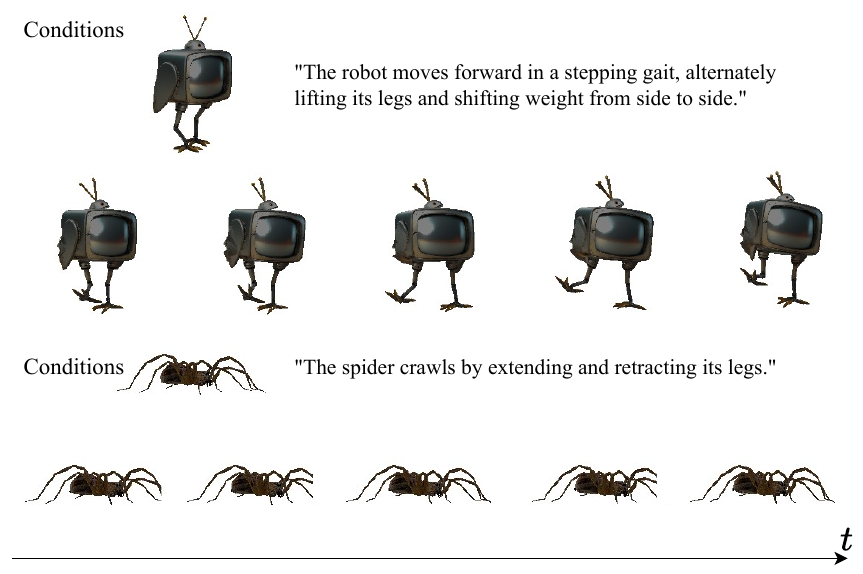}
\vspace{-15pt}
\caption{\textbf{Puppeteer Qualitative Results.} Results are rendered using the authors’ open-source renderer, whose texture appearance differs from ours and AnimateAnyMesh.}
\label{fig:puppeteer}
\end{figure}

\noindent\textbf{Puppeteer.} 
We also evaluate Puppeteer \cite{songpuppeteer}, a three-stage pipeline for driving the animation of a given mesh. First, it performs automatic rigging (skeleton and skinning estimation). Second, the static mesh is rendered to produce an initial conditioning frame for video generation. Third, it optimizes the  frame-by-frame motion by driving the estimated skeleton, so that the rendered frames match Kling-generated video conditioned on the rendering and prompt~\cite{kling2024}. 
As shown in \cref{fig:puppeteer}, Puppeteer is sensitive to rigging quality. 
For example, an incorrect rig can cause the Spider sequence to remain static. In addition, Puppeteer easily accumulates errors from early frames during the optimization process, leading to severe artifacts in some cases, such as the Robot example. 
In our experiments, Puppeteer consumed a peak GPU memory of approximately  48 GB and required about 58 minutes of runtime on a single A100 GPU to process a single sequence.

\begin{figure}[!htb]
\centering
\includegraphics[width=\linewidth]{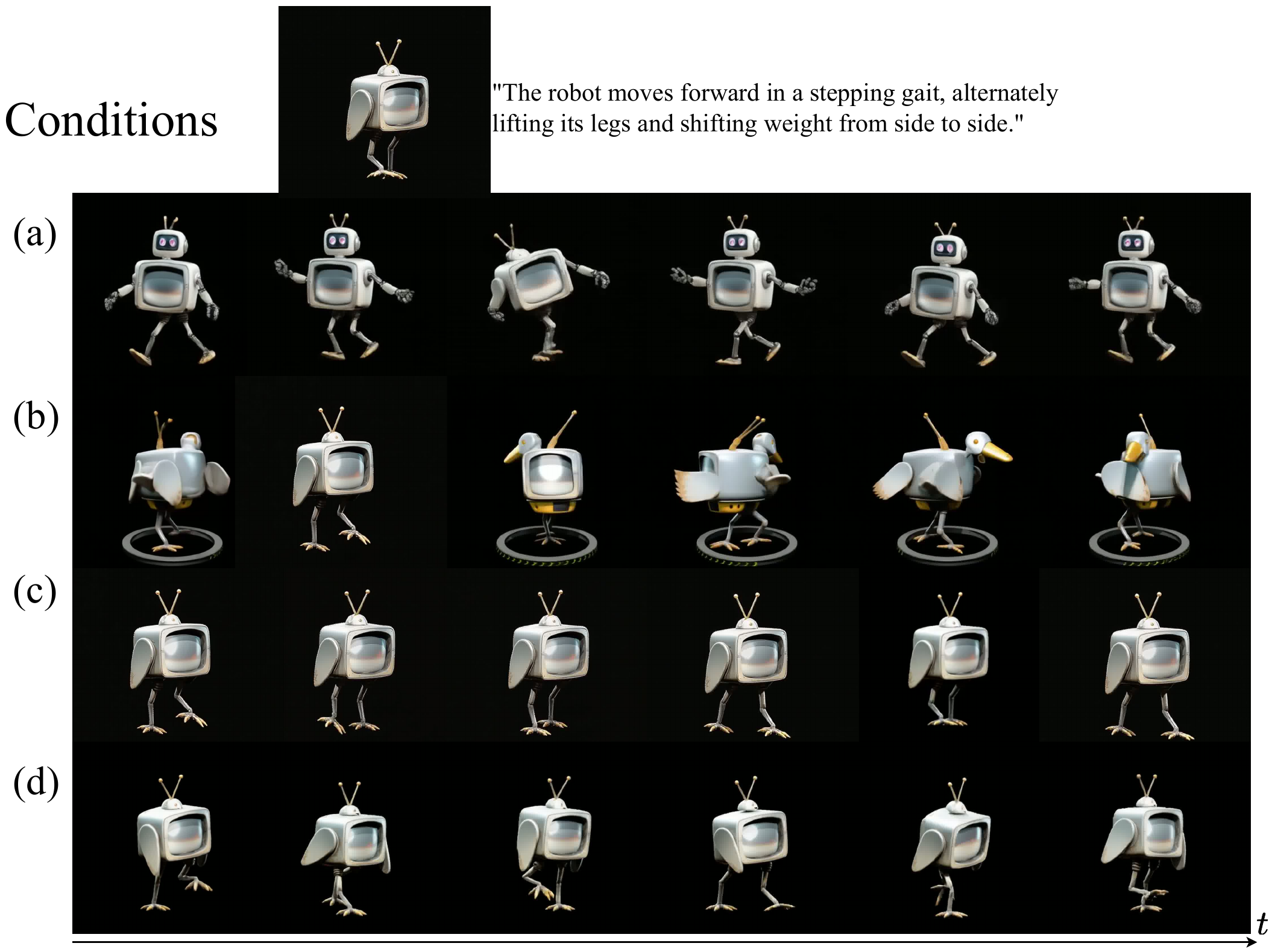}
\vspace{-10pt}
\caption{\textbf{Video Generation Model Comparison.} Qualitative comparison of video outputs from (a) Adobe Firefly \cite{adobe_firefly_video}, (b) HunyuanVideo-I2V \cite{kong2024hunyuanvideo}, (c) Wan-2.5 \cite{wan2025wan}, and (d) Kling~2.1 \cite{kling2024}. Note: these generation models convert transparent backgrounds to black by default.}
\label{fig:videogeneration_ablations}
\vspace{-4mm}
\end{figure}

\noindent\textbf{Video Generation Models.}  
For comparison with video-based dynamic 3D generation methods \cite{chen2025v2m4,zhang2025gaussian}, we use Kling~2.1 \cite{kling2024} as the video source, conditioned on the first frame and the text prompts. To validate this choice, we compare Kling with other state-of-the-art video generation models: Adobe Firefly \cite{adobe_firefly_video}, HunyuanVideo-I2V \cite{kong2024hunyuanvideo}, and Wan-2.5 \cite{wan2025wan}. As shown in \cref{fig:videogeneration_ablations}, Adobe Firefly and HunyuanVideo-I2V both struggle to maintain subject consistency and often introduce unrelated or unstable visual content. Wan-2.5 produces motion that aligns with the text prompts but lacks the expressiveness achieved by Kling.

\subsection{Additional Ablations} 
\noindent\textbf{Ablation of Number of Control Points.} We use 16 control points in all experiments to remain consistent with our baselines (\eg AnimateAnyMesh \cite{wu2025animateanymesh}). 
In new results in Tab.~\ref{tab:abl}, on the DeformingThings4D split, we increase the number of control points yielding trajectories closer to the raw motion during preprocessing. However, more control points necessitates larger training cost.

\begin{table}[!htb]
  \centering
  \caption{Ablation of the number of control points.}
  \resizebox{0.8\linewidth}{!}{
  \begin{tabular}{lcccc}
    \toprule
    Number of control points & 4 & 8 & 16 & 32\\
    \midrule
    L2 Error ($\times 10^{-3}$) $\downarrow$ &68.7 &22.6 & 6.85& 1.81\\
    \bottomrule
  \end{tabular}}
  \label{tab:abl}
\vspace{-5mm}
\end{table}

\begin{figure}[!htb]
\centering
\includegraphics[width=\linewidth]{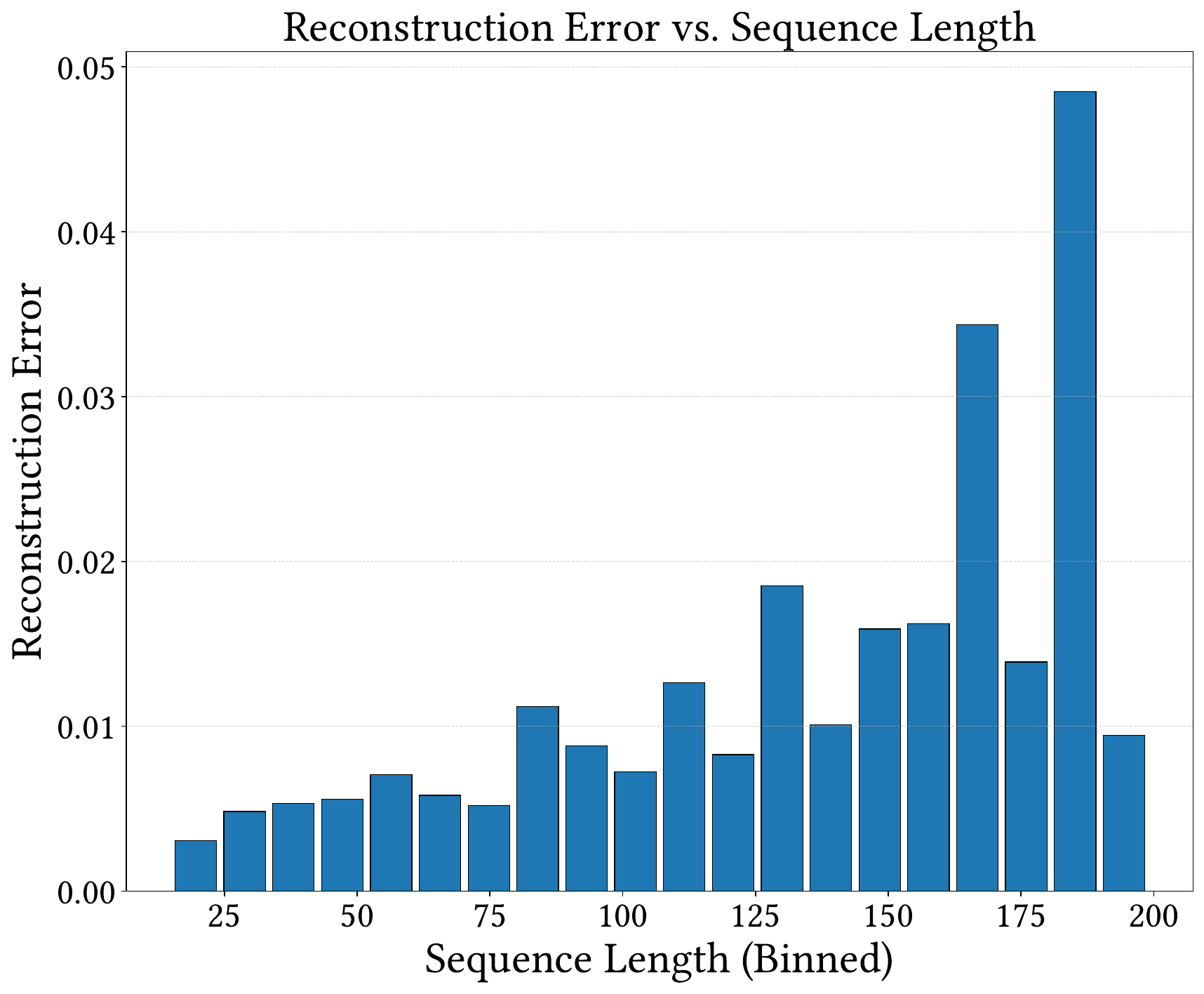}
\vspace{-15pt}
\caption{\textbf{VAE Reconstruction Error vs.~Sequence Length.} Mean L1 reconstruction error for our VAE, computed per trajectory and averaged per instance, reported for each sequence-length bin.}
\label{fig:addition_variable}
\end{figure}

\noindent\textbf{Variable-Length Sequence Representation.}
With \methodname, the B-spline-based VAE robustly handles variable-length sequences, even with a limited number of control points. 
In addition to the mean statistics reported in Tab.~\textcolor{myblue}{2} of the main paper, we provide binned reconstruction-error averages for a more fine-grained analysis. 
In \cref{fig:addition_variable} we report the mean L1 reconstruction error computed per trajectory and averaged per instance within each sequence-length bin. While our method reliably models short- and medium-length sequences, reconstruction error increases for longer sequences unless more than 16 control points are used. 
To ensure a fair comparison with AnimateAnyMesh, we fix the number of control points to 16 in all experiments. 

\begin{figure}[htb!] \centering 
\vspace{-4.2mm}
\includegraphics[width=1.\linewidth]{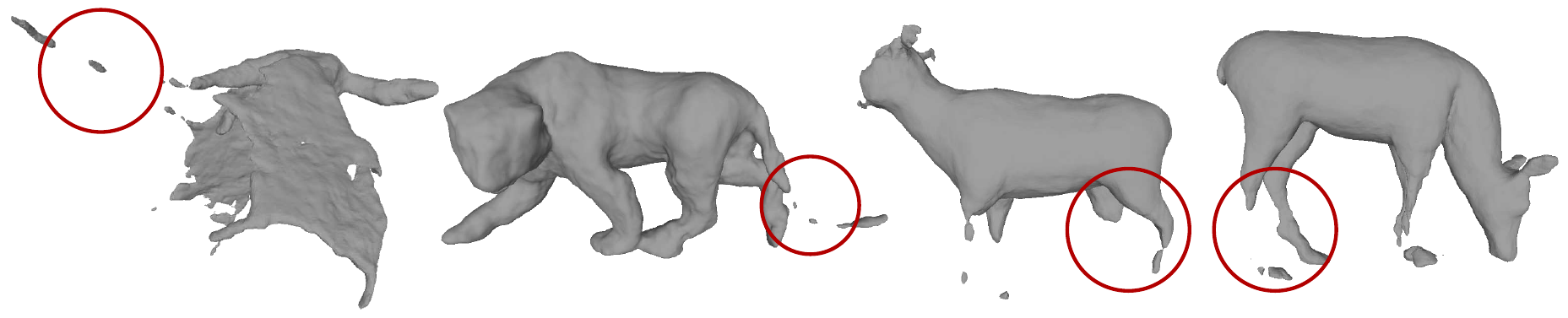}
\vspace{-8.5mm}
\caption{INR methods like  DNF can introduce shape artifacts.}
\vspace{-4mm}
\label{fig:implicit}
\end{figure}

\vspace{2mm}
\noindent\textbf{Implicit Neural Representations (INRs).}
Recent studies (CanFields \cite{CanFields2025}) show that 4D INRs are computationally expensive due to per-frame marching cubes extraction and also struggle to maintain consistent mesh identity over time. 
In Fig.~\ref{fig:implicit} we observe that the SOTA INR method DNF \cite{zhang2024dnfunconditional4dgeneration}, produces noticeable artifacts. 
In contrast, our task focuses on animating existing static meshes with fixed topology, which is better aligned with modern animation pipelines and allows consistent texture preservation without re-mapping.

\noindent\textbf{Additional Normal Fusion Ablations.}
We evaluate two alternative designs for incorporating surface normals into the initial shape features. 
(1) \textbf{$\mathcal{N}_0$ Frequency-PE.} Normals \(\mathcal{N}\) are encoded using the same frequency positional encoding (Frequency-PE) applied to the point coordinates~\cite{mildenhall2021nerf} and then passed through an MLP to obtain \(\mathbf{F}_{\mathcal{N}_0}\in\mathbb{R}^{n\times c}\). 
The fused feature \(\mathbf{F}_0\) is computed using Eq.~\textcolor{myblue}{5}.  
(2) \textbf{$\mathcal{N}_0$ Concat.} As in the main paper, we first extract \(\mathbf{F}_{\mathcal{N}_0}\) from \(\mathcal{N}\) using an MLP. To fuse normals, we concatenate \(\mathbf{F}_{\mathcal{N}_0}\) with \(\mathbf{F}_{\mathcal{P}_0}\) along the feature dimension and feed the result into another MLP to obtain \(\mathbf{F}_0\).

We report the VAE reconstruction error for these variants in \cref{tab:normal_abl}. Encoding normals with Frequency-PE disrupts their spherical geometric structure, introducing noise and degrading reconstruction accuracy. In contrast, our cross-similarity fusion aligns the MLP-projected normal features to the same latent space as the 3D position features, enabling more effective fusion and yielding substantially improved reconstruction compared with naive concatenation.

\begin{table}[!htb]
  \centering
  \caption{\textbf{Additional normal-fusion ablations.} Mean L1 reconstruction error (\(\times 10^{-2}\)), averaged per trajectory and per instance.}
  \vspace{-5pt}
  \resizebox{0.8\linewidth}{!}{
  \begin{tabular}{cccc}
    \toprule
    w/o $\mathcal{N}_0$ & $\mathcal{N}_0$ Frequency-PE & $\mathcal{N}_0$ Concat. & \textbf{Ours} \\
    \midrule
    1.328 & 1.331 & 1.294 & \textbf{1.078} \\
    \bottomrule
  \end{tabular}}
  \label{tab:normal_abl}
\end{table}

\subsection{User Study Details} 
We conducted an anonymous user study %via Google Forms 
to evaluate the perceptual quality of text-driven mesh animations. 
A total of 20 participants with diverse levels of experience in computer vision and graphics were recruited. 
Each participant evaluated four generation methods on the same ten randomly selected example sequences and rated three criteria on a 5-point Likert scale (1 = Very Poor, 5 = Excellent). 
We evaluated the following three questions for each sequence: (i) \textit{Text-to-Motion Agreement} (TA), (ii) \textit{Motion Plausibility} (MP), and (iii) \textit{Motion Expressiveness} (ME). In total, each participant contributed $4 \times 10 \times 3 = 120$ ratings.

\begin{figure*}[!htb]
\centering
\includegraphics[width=\linewidth]{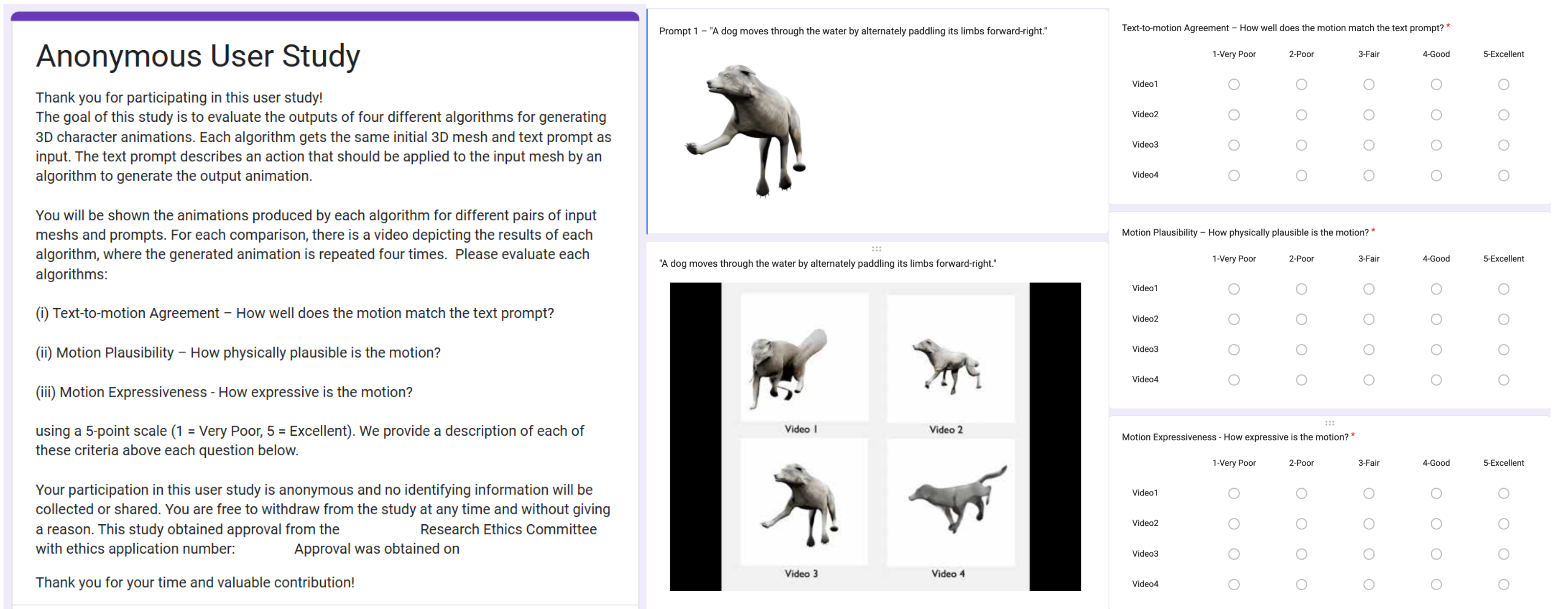}
\caption{\textbf{User Study Interface and Protocol.} 
\textbf{Left:} Instructions page. Screenshot of the questionnaire introduction; personal or institutional information approved by the research ethics committee has been masked. 
\textbf{Middle:} Video comparison interface showing the conditions of both the first frame and the textual prompt, along with four videos generated by four randomly ordered methods.
\textbf{Right:} Three evaluation questions used to collect user preferences, forming the basis of the subjective evaluation metrics.}
\label{fig:interface_protocol}
\end{figure*}

\noindent\textbf{Interface and Protocol.} As shown in \cref{fig:interface_protocol}, for each example object, the form displayed a 2$\times$2 grid of videos labeled Video~1–4, each corresponding to one of the four different methods.
To avoid bias, the mapping from methods to Video~1–4 was anonymized and randomized per participant and per example. The physical positions of the four videos in the 2$\times$2 grid were also randomly permuted for every page load. Participants could replay the clips freely before submitting their scores for TA/MP/ME (separate questions immediately under the grid). All responses were collected anonymously; participants could withdraw at any time without providing a reason.  
This study obtained ethics approval from the relevant internal ethics review board.  

\noindent\textbf{Results.} 
We report both the mean and standard deviation for each method in \cref{tab:user_std_model}. 
There we observe that users consistently favor \methodname, which achieves the highest mean score and the lowest standard deviation across all metrics. For detailed data, please refer to the raw survey metadata in the accompanying supplementary material, titled ``Anonymous User Study (Responses) – Form responses 1.csv''.
%Detailed per-user mean scores are provided in \cref{table:users_details}.

\definecolor{LightCyan}{rgb}{0.88,1,1}
\begin{table}[!htb]
\centering
\caption{\textbf{User study results.} Each cell shows the mean $\pm$ standard deviation over all participants and examples.} 
 \vspace{-5pt}
\label{tab:user_std_model}
\resizebox{\linewidth}{!}{
\begin{tabular}{l ccc}
\toprule
Methods & TA $\uparrow$ & MP $\uparrow$ & \multicolumn{1}{c}{ME $\uparrow$}\\
\midrule
GVFDiffusion \cite{zhang2025gaussian} & $2.343^{\pm{0.407}}$ & $2.300^{\pm{0.380}}$ & $2.443^{\pm{0.387}}$ \\
AnimateAnyMesh \cite{wu2025animateanymesh} & $2.314^{\pm{0.394}}$ & $2.686^{\pm{0.464}}$ & $2.443^{\pm{0.357}}$ \\
V2M4 \cite{chen2025v2m4} & $2.876^{\pm{0.368}}$ & $2.714^{\pm{0.354}}$ & $3.048^{\pm{0.348}}$  \\
\rowcolor{LightCyan}
\textbf{\methodname} (Ours) & $\mathbf{4.095^{\pm{0.334}}}$ & $\mathbf{4.062^{\pm{0.328}}}$ & $\mathbf{4.048^{\pm{0.339}}}$ \\
\bottomrule
\end{tabular}
}
\end{table}

\begin{figure*}[!t]
    \centering
    \includegraphics[width=\linewidth]{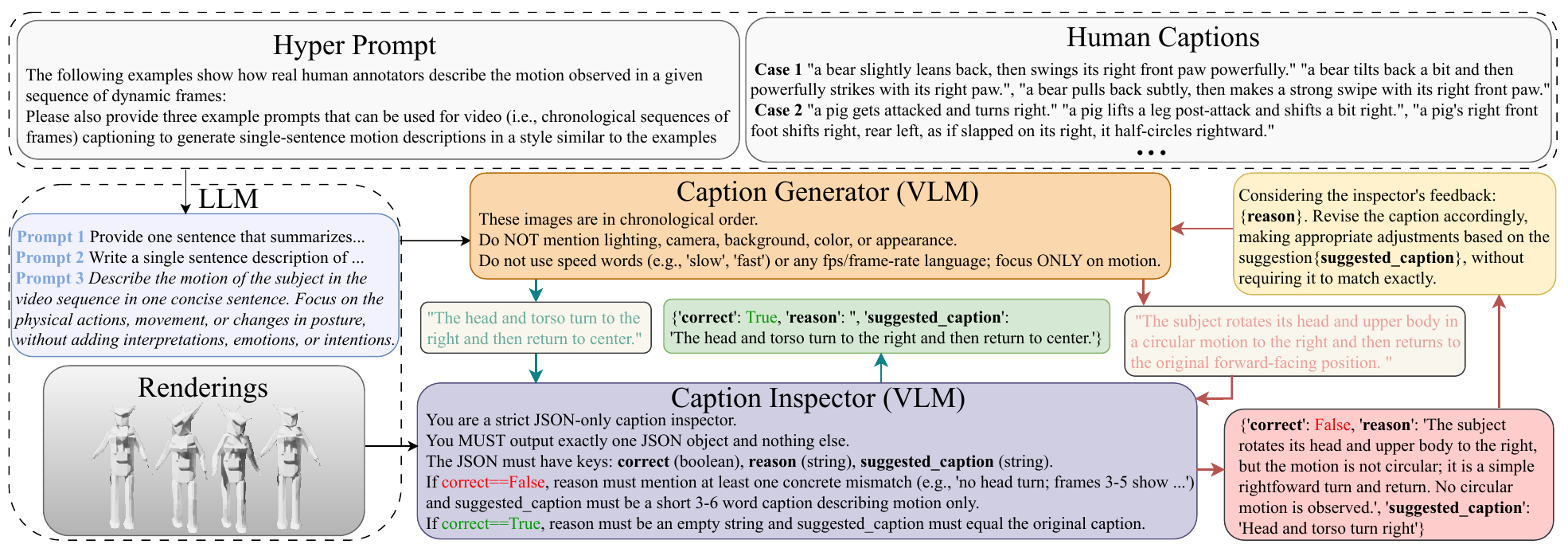}
    \vspace{-10pt}
\caption{\textbf{Auto-captioning pipeline.} We propose an iterative captioning pipeline that incorporates both a caption inspector and a caption generator to produce accurate and consistent captions for dynamic assets.}
    \label{fig:auto_caption}
\end{figure*}
\section{Dataset and Preprocessing}
\label{sec:dataset}
\subsection{Dataset Curation}
\begin{figure}[!htb]
    \centering
\includegraphics[width=\linewidth]{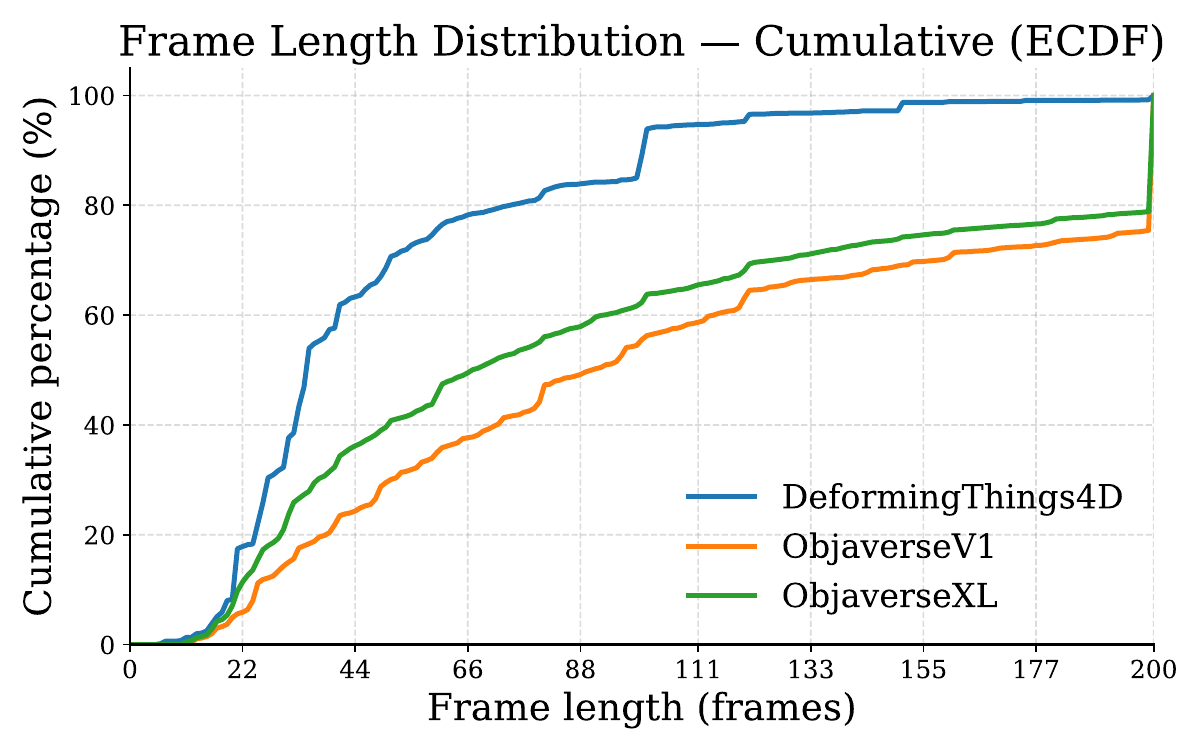} 
    \vspace{-20pt}
    \caption{\textbf{Frame-length cumulative distribution.} Distribution of mesh sequence lengths (length $\leq 200$) from DeformingThings4D~\cite{li20214dcomplete}, ObjaverseV1~\cite{deitke2023objaverse}, and ObjaverseXL~\cite{deitke2023objaverseXL}.}
    \label{fig:frame_length_cdf}
\end{figure}

To validate our method, we curated \textbf{BIMO} (Bspline Motion) dataset. We collect diverse mesh motion sequences from multiple datasets: DeformingThings4D~\cite{li20214dcomplete}, ObjaverseV1~\cite{deitke2023objaverse}, and ObjaverseXL~\cite{deitke2023objaverseXL}. For DeformingThings4D, we use the official script to extract animal mesh sequences from the raw `.anime' files. For ObjaverseV1 and ObjaverseXL, we %develop a Blender script to 
extract sequences with more than ten frames, discarding low-motion sequences based on object-center displacement, and bounding-box scale changes. 
We handle multiple raw formats from Objaverse (\eg ``.glb'', ``.gltf'', ``.fbx'') and filter out unnecessary elements such as ``icosphere'', ``sphere'', ``proxy'', ``env'', ``origin'', ``pivot'', ``camera'', ``ground'', ``floor'', and ``light''. The ObjaverseV1 asset IDs are curated by Diffusion4D~\cite{liang2024diffusion4d}, while ObjaverseXL meta IDs are filtered by GVFDiffusion~\cite{zhang2025gaussian}. All mesh sequences are capped at 200 frames following AnimateAnyMesh~\cite{wu2025animateanymesh}. The distribution of mesh sequence lengths is illustrated in \cref{fig:frame_length_cdf}, and the summary statistics of our curated dataset are presented in \cref{tab:dataset_stats}. 
The final dataset comprises 38,944 motion sequences with a total of 3,682,790 frames.

%------------------- Figure -------------------

%------------------- Table -------------------
\begin{table}[!htb]
    \caption{\textbf{Dataset statistics.} Summary statistics of our curated mesh motion dataset \textbf{BIMO}.}
    \vspace{-5pt}
    \centering
    \resizebox{1\linewidth}{!}{
        \begin{tabular}{l r r}
            \toprule
            Dataset Source & Number of Motions & Total Frames \\
            \midrule
            DeformingThings4D~\cite{li20214dcomplete} & 1,770 & 87,052 \\ObjaverseV1~\cite{deitke2023objaverse}      & 10,550 &  1,116,134 \\ ObjaverseXL~\cite{deitke2023objaverseXL}& 26,624 & 2,479,604 \\
            \midrule
            Total motions     & 38,944 & 3,682,790 \\
            \bottomrule
        \end{tabular}
    }
    \label{tab:dataset_stats}
\end{table}

\subsection{Auto-Captioning Pipeline}

To obtain captions for conditioning our model, we use human-annotated motion descriptions provided by OmniMotionGPT \cite{yang2024omnimotiongpt} for DeformingThings4D, where each motion is annotated with three captions written by trained annotators. 
For Objaverse, there are no captions available, and so we build an automatic captioning pipeline (see \cref{fig:auto_caption}) designed to emulate the style of the DeformingThings4D annotations. 
To reduce LLM token usage, we adopt a hyper-prompt design that first summarizes human annotations with an LLM to produce three concise prompts for a VLM. 
These distilled prompts, paired with the rendered frames, are supplied to the VLM as an alternative to requiring expensive to obtain human annotations for each caption. 
Concretely, we prompt an LLM (GPT-5~\cite{openai2025gpt5}) to imitate human annotators and generate three distinct video-captioning prompts. 
Our initial hyper-prompt that is provided to the LLM is:
\begin{tcolorbox}[colback=gray!5!white, colframe=gray!80!black, boxrule=0.5pt, arc=4pt, title=Hyper Prompt, breakable]
The following examples show how real human annotators describe the motion observed in a given sequence of dynamic frames:
\begin{description}
    \item[Case 1] A bear stands up and then returns to all fours. Another description states that the bear pushes with its front paws, stands on its hind legs, and then settles back down. A third annotator writes that the bear rose to its feet, used its front paws for support, stood upright, and then rested.
    \item[Case 2] A bull makes a left turn while remaining in one spot in the water. Another description notes that the bull turns in place, while a third says the bull swims in circles in the water.
    \item[Case 3] A wolf retreats in fright. Another annotator writes that the wolf withdraws in fear, while a third notes that the wolf is startled on the left and retreats to the right.
    \item[Case 4] A wolf falls into the water and moves forward. Another version describes it as dropping into the water, while a third says it plunges into the water and relaxes.
    \item[Case 5] A deer falls from above downward. Another description states that the deer descends from a higher to a lower position, while a third says it drops from high to low.
    \item[Case 6] A deer swims to the left while staying in place. Another annotator describes it as paddling left without advancing, while a third writes that it moves left in the water without forward motion.
\end{description}

Please also provide three example prompts that can be used for video (\ie chronological sequences of frames) captioning to generate single-sentence motion descriptions in a style similar to the examples above.
\end{tcolorbox} 

Given the above hyper-prompt, we obtain the following three auto-captioning prompts from the LLM which concisely summarizes the original hyper-prompt:

\begin{tcolorbox}[colback=blue!5!white, colframe=blue!80!black, boxrule=0.5pt, arc=4pt, title=LLM Generated Prompt,breakable]
\begin{enumerate}
    \item Describe the motion of the subject in the video sequence in one concise sentence. Focus on the physical actions, movement, or changes in posture, without adding interpretations, emotions, or intentions.
    \item Write a single-sentence description of what the subject does in the given sequence of frames. Include details about direction, movement, or interactions with the environment, but remain objective and observable.
    \item Provide one sentence that summarizes the main motion occurring in the video. Describe only what is visually happening, such as movement, turning, rising, falling, or swimming, in chronological order.
\end{enumerate}
\end{tcolorbox}

Next, we render the Objaverse dynamic assets from a fixed front-view camera using a modified Blender script adapted from Diffusion4D \cite{liang2024diffusion4d}. Rendering is performed on ten A6000 GPUs with the Cycles engine, producing transparent 256×256 frames. We then use a VLM (GPT-5) to generate captions for each rendered sequence. To ensure consistency and reliability, we incorporate an iterative caption-inspection loop using the same model. 
The caption inspector (implemented with the same VLM GPT-5) receives both the renderings and the captions produced by the caption generator, evaluates their correctness, and provides reasoning and suggested revisions when necessary. The caption-generation agent then updates the caption and repeats  the process six times until the inspector approves. 

\subsection{VLM Ablations}

\begin{figure}[!htb]
\centering
\includegraphics[width=\linewidth]{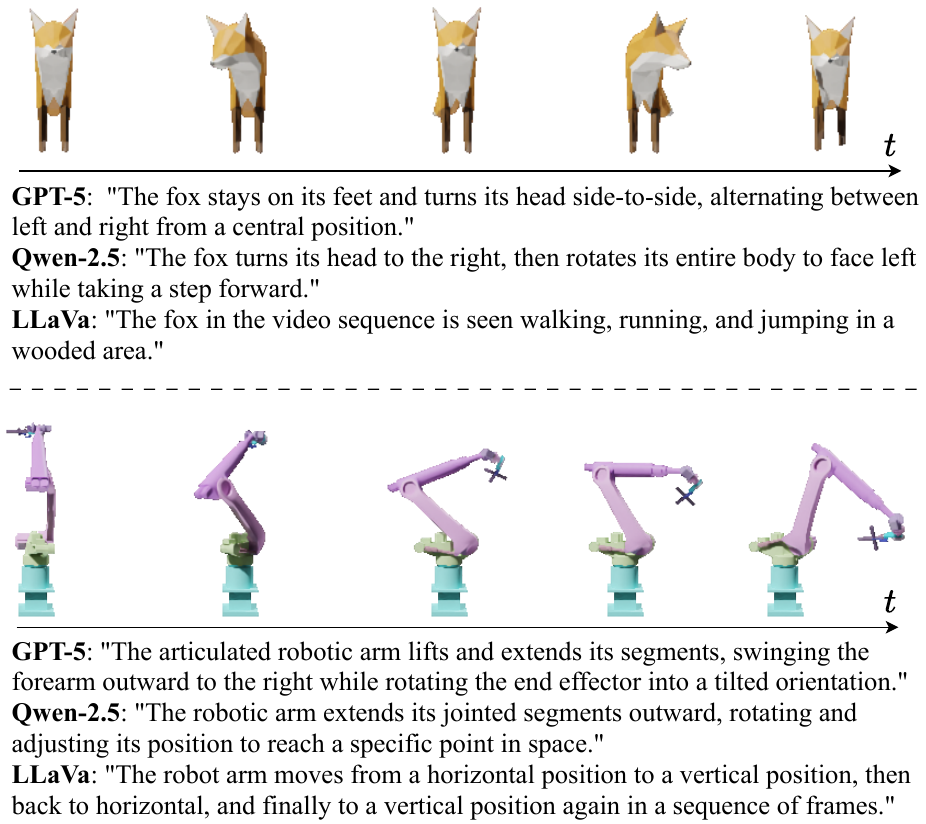}
\vspace{-15pt}
\caption{\textbf{VLM Captioning Ablation.} 
We compare captions generated by three video–language models (VLMs). 
Using GPT-5 produces concise and motion-faithful descriptions closely aligned with the real 4D dynamics.}
\label{fig:vlm_ablations}
\end{figure}

To validate our VLM choice, we ablate GPT-5, Qwen-2.5~\cite{qwen2.5-2024}, and LLaVA-NeXT-Video~\cite{liu2024llavavideo} in \cref{fig:vlm_ablations}. GPT-5 consistently yields motion-focused, non-speculative descriptions that match the DeformingThings4D annotation style. Qwen-2.5 frequently hallucinates global behaviors (\eg stepping forward, full-body reorientation) and conflates pose with inferred intent. LLaVA-NeXT-Video more often attributes locomotion or complex actions absent from the frames. These errors shift the caption distribution away from true 4D motion and weaken text–motion correspondence during training. Our GPT-5 hyper-prompt pipeline produces stable, grounded, and fine-grained captions that improve the robustness and fidelity of downstream text-conditioned animation synthesis.

% \clearpage
% \newpage
% {
%     \small
%     \bibliographystyle{ieeenat_fullname}
%     \bibliography{main}
% }